%% file: neurips_2025.tex
\documentclass[11pt, letterpaper]{berkeley}

\usepackage[utf8]{inputenc} %
\usepackage[T1]{fontenc}    %
\usepackage{hyperref}       %
\usepackage{url}            %
\usepackage{booktabs}       %
\usepackage{amsfonts}       %
\usepackage{nicefrac}       %
\usepackage{microtype}      %
\usepackage{xcolor}         %
\usepackage{algpseudocode}
\usepackage{algorithm}
\usepackage{enumitem}

\usepackage{amsmath}
\usepackage{amssymb}
\usepackage{mathtools}
\usepackage{amsthm}
\usepackage[frozencache,cachedir=.]{minted}
\usepackage{listings}
\usepackage{makecell}
\usepackage{tabularx}
\usepackage{lscape}
\usepackage{wrapfig}
\usepackage[capitalize,noabbrev]{cleveref}
\input{math_commands}

\def\ab{\text{AB}}
\def\abclip{\ab$_{\text{CLIP}}$}
\def\absimclr{\ab$_{\text{SimCLR}}$}
\def\abdino{\ab$_{\text{DINO}}$}
\def\bb{\mathbf{bb}}

\newcommand{\Th}[1]{\small\textsc{#1}}

\def\ema{\textnormal{ema}}
\newcolumntype{Y}{>{\centering\arraybackslash}X}
\newcolumntype{C}{>{\centering\arraybackslash}X}

\title{Visual Pre-Training on Unlabeled Images \\ using Reinforcement Learning}
\runningtitle{Visual Pre-Training on Unlabeled Images using Reinforcement Learning}

\reportnumber{} %

\author[ ]{Dibya Ghosh}
\author[ ]{Sergey Levine}

\affil[ ]{UC Berkeley}

\correspondingauthor{Dibya Ghosh (\href{mailto:dibya@berkeley.edu}{dibya@berkeley.edu})}

\title{Visual Pre-Training on Unlabeled Images \\ using Reinforcement Learning}

\begin{document}

\begin{abstract}

 In reinforcement learning (RL), value-based algorithms learn to associate each observation with the states and rewards that are likely to be reached from it. We observe that many self-supervised image pre-training methods bear similarity to this formulation: learning features that associate crops of images with those of ``nearby'' views, e.g., by taking a different crop or color augmentation. In this paper, we complete this analogy and explore a method that directly casts pre-training on unlabeled image data like web crawls and video frames as an RL problem. We train a general value function in a dynamical system where an agent transforms an image by changing the view or adding image augmentations. Learning in this way resembles crop-consistency self-supervision, but through the reward function, offers a simple lever to shape feature learning using curated images or weakly labeled captions when they exist. Our experiments demonstrate improved representations when training on unlabeled images in the wild, including video data like EpicKitchens, scene data like COCO, and web-crawl data like CC12M.

\end{abstract}

\maketitle

\section{Introduction}

\begin{wrapfigure}{r}{0.4\linewidth}\centering
    \includegraphics[width=0.9\linewidth]{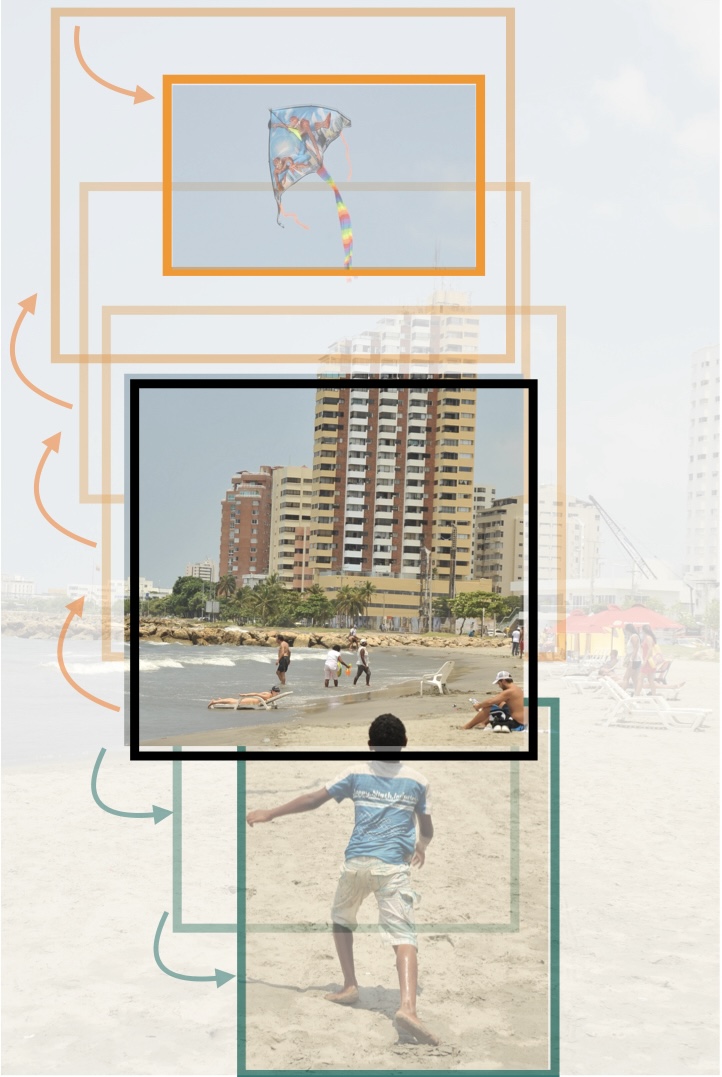}
\caption{In annotation bootstrapping, we train on unlabeled images with RL: predicting and maximizing semantic annotations associated with distant sub-crops of an image (e.g., a kite, or a boy playing). Our key idea is that we can learn this from unlabeled images, even though they provide no supervision of their own. We instead learn by \textit{bootstrapping} between image crops: using model outputs on one view to provide a target for our model on a nearby crop.}
\label{fig:intro}
 \vspace{-1.5em}
\end{wrapfigure}
Self-supervised representation learning is one of the basic building blocks of modern large-scale machine learning. From next-token prediction in NLP to contrastive representation learning in computer vision, methods that can leverage large unlabeled corpora (e.g., mined from the web) can acquire powerful representations that can then be used to efficiently solve downstream tasks. 

However, representation learning for highly unstructured modalities such as images remains very challenging, as raw pixels offer a learning signal with redundancy, low information density, and noise. To learn useful features, methods often rely on weak supervision like textual captions \citep{radford2021learning, Yu2022CoCaCC, Zhai2023SigmoidLF} or carefully designed inductive biases like feature invariance to image augmentations and crops \citep{He2019MomentumCF, Chen2020ASF, Caron2021EmergingPI}. It is unclear how to generalize these inductive biases to broader unlabeled image datasets, like web crawl or video data, or other downstream tasks like object detection or embodied action recognition.

In this paper, we make the observation that unlabeled images can be associated with annotated images through ``long horizon'' relationships that involve a series of random image transformations. Taking inspiration from temporal difference (TD) learning in reinforcement learning (RL), we notice that data augmentations (e.g., random crops) determine a kind of ``transition dynamics,'' while annotated images provide ``rewards'' in a sense. Just as a TD algorithm in RL propagates sparse reward information in an environment by bootstrapping from one state to the previous, we may iteratively propagate annotation information from image views to nearby crops to learn across unlabeled data.

We make this analogy explicit and consider directly using a TD-based value learning algorithm to pre-train visual features from unlabeled image data. This results in a method we call \textit{annotation bootstrapping}, which learns a general value function \citep{Sutton2011HordeAS} for an environment where actions apply controllable automated image transformations like random crops, and rewards measure the likelihood of eliciting specific semantic annotations (e.g., ``a kite'' or ``a boy playing'' in Figure \ref{fig:intro}).

By choosing what reward signals the value function will model over the transformation process, we can focus representation learning toward specific semantic concepts of interest. For instance, defining rewards using captions from weakly-labeled data results in value functions that train by iteratively predicting distributions over captions associated with neighboring crops of an image. In experiments using CC-12M \citep{Changpinyo2021Conceptual1P}, we found bootstrapping trains better representations than methods that directly combine weak supervision and standard self-supervised losses. 

Rewards may also be defined by unlabeled images. For example, we can learn a reward function by maximizing crop-consistency on a curated dataset, and train a value function to model these rewards on other unlabeled image datasets where this inductive bias may not fit. In our experiments, we show that on COCO \citep{Lin2014MicrosoftCC} and Epic-Kitchens \citep{Damen2020TheED}, whose images are not object-centric, value learning from crop-consistency rewards yields better features than directly maximizing crop-consistency on these images.

Our main contribution is annotation bootstrapping, a framework for self-supervised representation learning that frames the learning problem as an RL task. By maximizing a value estimation objective, the model learns to jointly understand relationships between images and semantic annotations, and between neighboring crops of images. This RL-based method can be applied on top of a variety of prior frameworks, and provides a surprisingly strong method for learning features from unlabeled images.
Beyond performance, framing representation learning as an RL task also offers useful perspectives on feature learning. For example, the reward function offers a simple way to guide learning with curated or supervised data, without needing to tune the masking or data augmentation strategy. Furthermore, while losses and model outputs for other self-distillation approaches often are uninterpretable, since we are learning a value function, model predictions can be inspected and analyzed throughout the learning process. Our experiments verify the effectiveness of annotation bootstrapping, especially for ``in-the-wild'' unlabeled datasets like web crawls and video frames where crop-consistency methods typically falter.

\section{Related Work}

\paragraph{Self-supervised learning.} Self-supervised methods generally learn in one of two ways: by reconstruction or enforcing representational consistency. Reconstruction-based approaches directly predict raw pixels \citep{He2021MaskedAA} or other low-level features \citep{Xie2021SimMIMAS, Bao2021BEiTBP} from masked or corrupted inputs. Although simple and easily scalable \citep{elnouby2024scalablepretraininglargeautoregressive, bai2024sequential, chameleonteam2024chameleonmixedmodalearlyfusionfoundation}. these objectives yield poor representations for downstream tasks, requiring finetuning and often strongly influenced by data curation \citep{elnouby2024scalablepretraininglargeautoregressive}.

Consistency-based approaches use carefully crafted objectives to learn better semantic features, most common being to enforce invariance under random crops and augmentations. Consistency can be optimized directly by contrastively attracting representations of paired views and repelling negative pairs \citep{Oord2018RepresentationLW, He2019MomentumCF, Chen2020BigSM, tian2020makes}, e.g., SimCLR \citep{Chen2020ASF}. Other approaches implicitly optimize for consistency by iterative self-distillation, e.g., DINO \citep{Caron2021EmergingPI} or BYOL \citep{Grill2020BootstrapYO}. These classes come with different challenges: contrastive methods are stable but require a large batch size to learn effectively \citep{He2019MomentumCF, Chen2021AnES}; self-distillation methods are more performant, but unstable and require careful architectural or objective changes, such as logit sharpening \citep{Caron2021EmergingPI}, k-means clustering \citep{Caron2019UnsupervisedPO}, non-differentiable transports \citep{Caron2020UnsupervisedLO}, or asymmetric predictors \citep{Grill2020BootstrapYO, Xie2021SimMIMAS}. 
Value bootstrapping is mechanistically similar to self-distillation, but has two benefits to crop-consistency. First, since value functions optimize for equivariance instead of invariance, they are less sensitive to the exact augmentation strategy \citep{Chen2020ASF, chenintriguing}, and generalize better to unlabeled data where invariance to random crops is a poor bias \citep{haochen2023theoreticalstudyinductivebiases, venkataramanan2023imagenet, jha2024commonstabilitymechanismselfsupervised}. Second, value learning has a stable fixed-point solution, while self-distillation methods can
suffer from representation collapse due to degeneracy in the loss landscape \citep{SimSiamCollapse, Jing2021UnderstandingDC}.

\paragraph{Vision-language pre-training.} Methods have found success in combining weakly-supervised learning, which learn to associate images and textual captions scraped from the internet \citep{radford2021learning, Jia2021ScalingUV, Zhai2023SigmoidLF}, with the self-supervised objectives above. SLIP \citep{Mu2021SLIPSM} combines CLIP with a SimCLR objective using an auxiliary head, \citet{Li2021SupervisionEE} jointly runs CLIP and SimCLR both on the same representation, and SiLC \citep{naeem2023silc} combines SigLIP with a DINO objective. Combining these losses improves the data efficiency of contrastive vision-language training, and improves performance for more fine-grained tasks like segmentation and prediction \citep{naeem2023silc}. One challenge with these approaches is appropriately weighting these two losses \citep{fini2023improved}; since both the supervised and unsupervised losses in annotation bootstrapping predict distributions over textual captions, combining the losses offers a more natural alignment. In our experiments, we find bootstrapping leads to better performance, and measuring gradient similarity, also aligns better with the weakly supervised loss throughout training.

\paragraph{Semi-supervised learning.} While our paper focuses on unlabeled image pre-training guided by descriptive annotations like free-form text, it is adjacent and inspired by a longer line of semi-supervised approaches learning with partially annotated class labels. Two techniques are common: combining a supervised classification loss with a self-supervised objective on unlabeled data \citep{Pathak2016ContextEF,Chen2020BigSM, Zhai2019S4LSS,Xie2019UnsupervisedDA}, and using the supervised dataset to create pseudo-labels \citep{lee2013pseudo} for unlabeled images \citep{Xie2019UnsupervisedDA, Pham2020MetaPL}. Both pseudo-labeling and annotation bootstrapping generate targets using model outputs, but with one important difference: pseudolabeling creates labels for a different student model for the same image, while value bootstrapping generates supervision for the {same model, but a different image}.

\section{Reinforcement Learning \textit{on} Images}

The core of our approach is to cast learning from unlabeled image datasets as a reinforcement learning problem. 
Training with RL will consist of two parallel threads: a \textit{reward} loss that associates images to annotations (useful semantic concepts), and a \textit{value} loss that associates crops of images with neighboring crops, to propagate these learned semantic relationships through the unlabeled images. 

\subsection{Annotation Bootstrapping}

We define an RL problem over unlabeled images using a Markov chain perspective on image augmentations \citep{Johnson2022ContrastiveLC}. An agent receives a view of an image $x$ and takes actions by applying an image transformation to change the view (e.g., zooming out, panning left, rotating the image, cropping to a subview), with the intention of finding views that maximize the likelihood of some specific semantic ``annotation'' of interest $p(\ell | x)$ (e.g., find the gentleman in the green suit, or a kite, or a boy playing). This environment is stochastic from the perspective of the agent: given a close-up of a dog, zooming out may reveal e.g., grass, pavement, or even a dog bed.

Formally, we learn in a multi-task MDP  where an action $a$ applies a parameterized stochastic image transformation $x, a \mapsto x'$ induced by the unlabeled image dataset. Given an annotation $\ell$, the agent's task is to maximize the expected likelihood of eliciting an annotation $\ell$ along images from this sequential transformation process, corresponding to a reward function $r(x) = p(\ell|x)$ (see Appendix \ref{app:rl} for an overview of relevant RL concepts and pertinent derivations in this section).

Value-based algorithms solve the RL task by learning to estimate the value function of the optimal policy $Q^*(x, a, \ell) \equiv Q^{\pi^*}(x, a, \ell)$ from samples of transitions $(x, a, x')$. The general idea of these methods is that the optimal function satisfies a (unique) Bellman fixed point: 
\begin{equation}
Q^*(x, a, \ell) = \mathcal{T}Q^*(x, a, \ell) \equiv \E_{x' \sim P(\cdot | x, a)}[r(x', \ell) + \gamma \max_{a'}Q^*(x', a', \ell)].    
\end{equation}

Recursively optimizing the value function at $(x, a)$ to match the estimate generated by $x'$, called bootstrapping, guarantees the model will converge to the optimal function \citep{Sutton1998}.

Applying this recursion in the image-transformation MDP leads to an optimization procedure very similar to standard self-supervised methods. Given two random crops of an image, $x_1$ by cropping to bounding box $\bb_1$ and similarly $x_2$ to $\bb_2$, we can interpret $(x=x_1, a=\bb_{1\to2}, x'=x_2)$ as a transition in our environment (writing $\bb_{1 \to 2}$ as the relative coordinates of $\bb_2$ relative to $\bb_1$): applying a panning transformation from the view $x_1$ to create $x_2$. Optimizing the value recurrence corresponds to, for any $\ell$, using the model's outputs at one crop $x_2$ to generate a target prediction for the other crop $x_1$:

\begin{equation}
\min D(~Q_\ab(\ell | x_1, a=\bb_{1\to 2}), (1-\gamma)p(\ell | x_2)  + \gamma \max_{a'} Q^{target}(\ell | x_2,a'))  
\end{equation}

We term this process \textit{annotation bootstrapping}, as the primary learning mechanism is the bootstrap: using the model's predictions about annotations in one image $x_2$ as supervision to improve its predictions about a nearby crop $x_1$. This can lead to a synergistic cycle: as the model improves its ability to predict semantics about the scene around it, it acquires a better semantic understanding of its own scene, thereby improving the targets used to train the model in the future. 

\subsection{Practical Implementation}

We now describe a practical algorithm that jointly learns a reward function $p(\ell | 
 x)$ and optimizes the value learning objective from a dataset of unlabeled images $x \sim D_u$. Value-learning can use rewards induced by any contrastive objective, but for clarity of exposition, we will consider learning a reward function from a dataset $(x, \ell) \sim D_a$ using an InfoNCE objective with form:
 \begin{equation}
\label{eq:reward_logits}
r(x, \ell) = p(\ell | x) \propto  \exp \left(t * \phi(x)^\top \psi(\ell)\right)
\end{equation}

 For instance, if $\ell$ corresponds to a text caption from a web crawl dataset, the reward function is trained by CLIP (InfoNCE between images and captions) and the value function objective is to predict distributions over captions associated with related images. In our experiments, we study three versions of reward functions: \abclip~where the reward function is trained via CLIP and annotations correspond to text captions, \absimclr~atop a SimCLR reward objective where annotations correspond to augmented image views $\ell = \operatorname{augment}(\operatorname{randomcrop}(x))$, and \abdino~atop a DINO reward objective, where annotations correspond to prototype clusters.

\input{my_algorithm}

In all three versions, the image representation $\phi(x)$ is a Vision Transformer \citep{Dosovitskiy2020AnII}, pooled and projected to a shared embedding space. The annotation representation $\psi(\ell)$ depends on the format of annotations: for textual captions, we use a symmetrically implemented text transformer, while for image annotations, we add a head to the existing backbone, matching \citep{radford2021learning} and \citet{Chen2020ASF} respectively. The value function also takes the contrastive form:
\begin{equation}
\label{eq:ab_logits}
Q^{\ab}(x, a, \ell) = \sigma \left(t_\text{\tiny AB} * \phi_\text{\tiny AB}(x, a)^\top \psi_\text{\tiny AB}(\ell_i) + b_\text{\tiny AB}\right)
\end{equation}

The ``image-action" representation $\phi(x, a)$ is implemented as a lightweight decoder atop the image backbone, processing a set of action tokens and cross-attending to the visual embeddings. The annotation representation $\psi_\ab(\ell)$ is identical to that for the reward function, as an independent head atop the annotation backbone. A visual diagram of the model is provided in Figure \ref{fig:enter-label} in the Appendix.

During training, we sample unlabeled image data and generate $n$ random crops of each image $I$: $x_i, \mathbf{bb_i} = \operatorname{RandomCrop}(I)$. For any two views $i, j$ of the same image, we can interpret $x_j$ as the ``next state'' resulting from taking the action $\bb_{i\to j}$ from the current state $x_i$. For any set of annotations $\{\ell_i\}_{i=1}^m$, we train the model for one view $x_i$ to match the estimated values generated by the other view $x_j$: 
\begin{equation}
\label{eq:bootstrapping}
    \gL_\ab = D(Q(x_i, \bb_{i \to j}, \ell), (1-\gamma)p^{\ema}(\ell|x_j) + \gamma \max_{a'} Q^{\ema}(x_j, a', \ell))
\end{equation}
\begin{wrapfigure}{r}{0.6\linewidth}
    \centering
    \includegraphics[width=\linewidth]{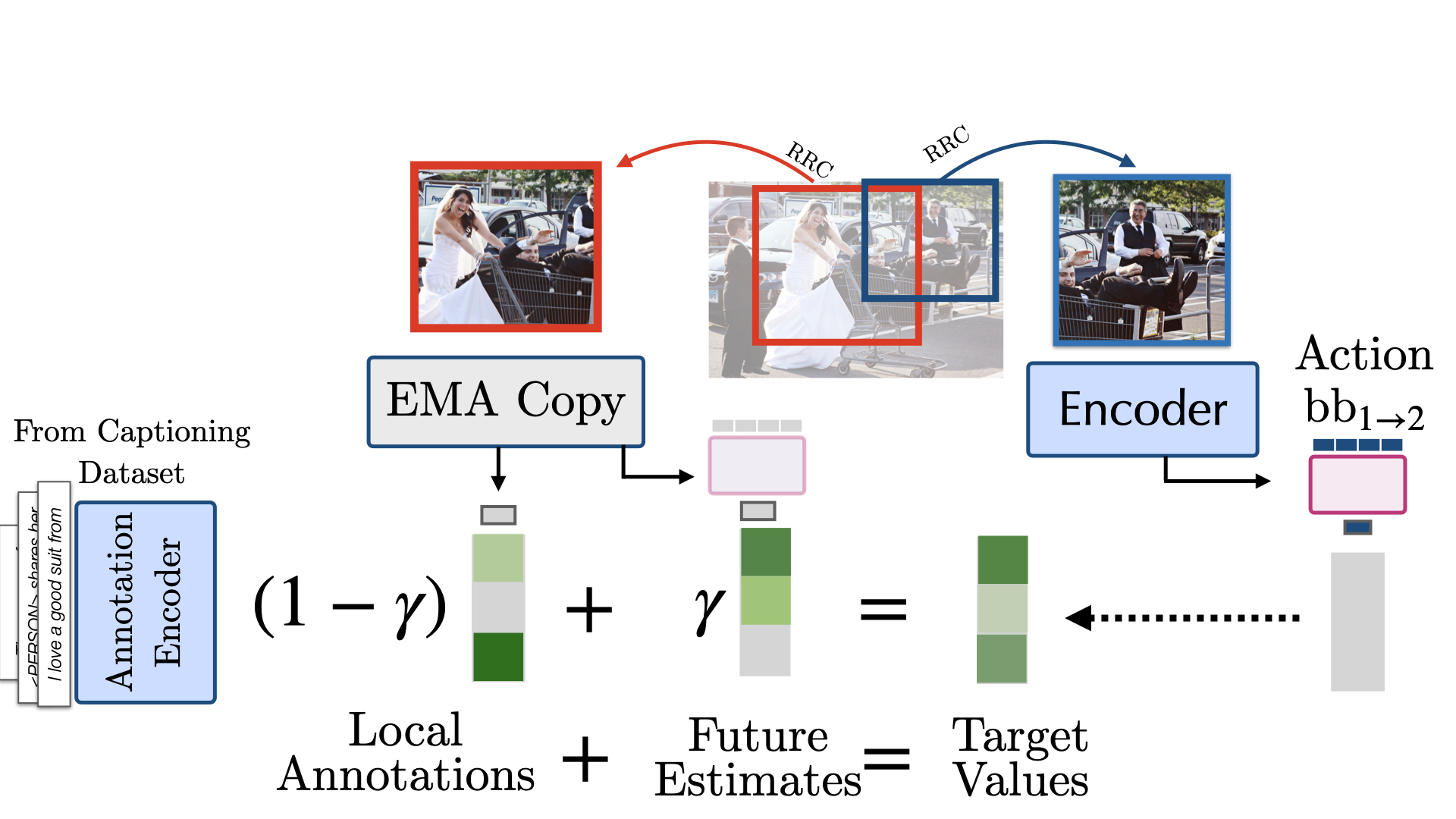}
    \caption{Visualization of the value function objective in our method with a base CLIP loss. The model processes a crop of an image $x_1$ and action tokens demarcating how to transform the image to a target crop $x_2$ (e.g. using the relative bounding box $\bb_{1 \to 2}$. The targets are created using the outputs of an EMA copy of the model on the view $x_2$.}
    \label{fig:bootstrapping}
\end{wrapfigure}
where the natural distance is the binary cross-entropy loss. As with other RL methods and self-distillation methods, we use a lagging EMA average of model parameters when computing the target distribution to ensure training stability. Through token packing and batching, this loss can be computed efficiently across all $n^2$ pairs of views. The overall algorithm is summarized in Algo. \ref{alg:mainalg} and Figure \ref{fig:bootstrapping}.

\subsection{Connections}

\textit{Soft distillation and pseudo-labeling.} The bootstrapping objective,
in form, resembles distillation objectives like pseudo-labeling \citep{Iscen2019LabelPF, semisupervised}, but induces a very different effect. Distillation transfers knowledge from one model $p_\theta$ to another $q_\theta$ about an image $x$; annotation bootstrapping instead transfers knowledge from one image $x_2$ to another $(x_1, \bb_{1\to 2})$, but for the same model. This distinction is significant, as we are interested in objectives that improve representation learning for the current model, not re-training a new model from scratch.

\textit{Consistency via self-distillation.} The bootstrapping objective is closely related to self-supervised methods that optimize for consistency via iterative self-distillation. Amongst others, DINO \citep{Caron2021EmergingPI} and SwAV \citep{Caron2020UnsupervisedLO} also predict distributions over ``prototypes'' (cf. annotations) associated with one crop $x_{2}$ from a different crop $x_1$. However, the DINO and SwAV objectives optimize for invariant representations, that crops of an image should emit the same distribution over prototypes. In contrast, annotation bootstrapping results in equivariant representations; in Figure 2, the red and blue should correspond to different annotation distributions since they capture different semantic details (like the wedding dress vs. a man in the background); value bootstrapping only attempts to make these distributions predictable from (not the same as) one another. 

Also similar is I-JEPA \citep{Assran2023SelfSupervisedLF}, which makes self-supervised predictions about the representation for a target view specified by positional tokens (c.f. action tokens in annotation bootstrapping) using a similar architecture. We note two high-level differences: I-JEPA predicts representations of states ``one transformation away'' in the system, while the recursion in the value objective develops associations with images further in the augmentation process. Second, the I-JEPA objective, predicting backbone tokens using an asymmetric predictor head \citep{Grill2020BootstrapYO, Chen2020ExploringSS} is not guaranteed to have a fixed solution, while the value prediction task is by construction has a guaranteed solution with well-studied convergence properties. 
We discuss lower-level implementation comparisons between these methods in Appendix \ref{app:connections}. 

\section{Experiments}
We study the utility of using annotation bootstrapping to learn representations from unlabeled images from a wide range of settings (web crawls, video data, scene data), and atop different ``reward functions'' of interest. Our study focuses on the following questions:

\begin{enumerate}[leftmargin=3em, itemsep=0mm]
    \item Can RL improve pre-training with different base reward losses like SimCLR, DINO, or CLIP?
    \item How does bootstrapping compare to invariance-based or pixel-predictive self-supervision?
    \item Can we bootstrap from curated datasets to learn better on a different unlabeled dataset?
\end{enumerate}

As we investigate these questions, we also probe the training process to understand how RL interfaces with the base reward loss, and the effect of various design decisions in value estimation. Full experimental details about the method, training, and evaluation are in Appendix \ref{app:train} and \ref{app:eval}. Code is provided at \url{github.com/dibyaghosh/annotation_bootstrapping}.

\textbf{Training.} We standardize training by running all methods on all datasets using ViT-S/16 vision encoders (and S-sized text encoders in the weakly labeled setting) for $800 M$ seen images (each view is counted separately), unless otherwise stated. For ImageNet, this corresponds to approximately $620$ epochs of the dataset. All models are trained with AdamW, weight decay, gradient clipping, and using a cosine decay schedule -- specific hyperparameters are taken from respective papers when they are provided (see Table \ref{table:hyperparameters} in Appendix \ref{app:train} for a full list).

We emphasize that our experimental goal is not to claim state-of-the-art performance on standard unsupervised benchmarks, but rather to evaluate RL-based representation learning on a wide set of domains and carefully analyze how it relates to patterns like crop-consistency and pixel reconstruction.

\input{tables/i1k_probe}

\textbf{Evaluation.} To avoid overfitting to Imagenet performance, we evaluate on a wider set of tasks using the probing strategy introduced by \citet{Beyer2023ASO}. In this setup, evaluation tasks (including classification, object detection, visual question answering, captioning, etc) are cast as autoregressive modeling tasks, and learned using a decoder that cross-attends with frozen ViT embeddings.  This solution allows us to evaluate a broad set of downstream tasks under a unified interface. %

\input{tables/cc12m}
\subsection{Pre-training with a self-supervised base loss}
We first evaluate annotation bootstrapping in the fully-unlabeled setting, where we use a base SimCLR (\absimclr) or DINO  (\abdino) loss to learn a reward signal, and train a value function to make predictions in the induced space of image-image relationships. 

When pre-training on unlabeled ImageNet images (Table \ref{table:i1k_probe}, additional probes in Table \ref{table:i1k_full}), a standard well-curated dataset, we find annotation bootstrapping to be synergistic to the base SimCLR / DINO loss, improving performance over running the base losses for a longer period of time. Investigating different probes of the visual representation, the improvement is greatest on probes that attend to the encoded tokens (like MAP pooling or a larger decoder), but not those that have been reduced to a single token (e.g. by global average pooling). These trends also hold when training using images without captions from CC-12M (Table~\ref{table:cc12m}), a larger and less curated dataset of web-crawl images common for vision-language training \citep{Changpinyo2021Conceptual1P}.

We find that crop-consistency methods significantly degrade when they are trained on non-object-centric data, specifically on COCO \citep{Lin2014MicrosoftCC} and Epic-Kitchens \citep{Damen2020TheED}, treating video data as individual frames. 

\begin{wrapfigure}[22]{r}{0.5\textwidth}
    \centering
    \includegraphics[width=\linewidth]{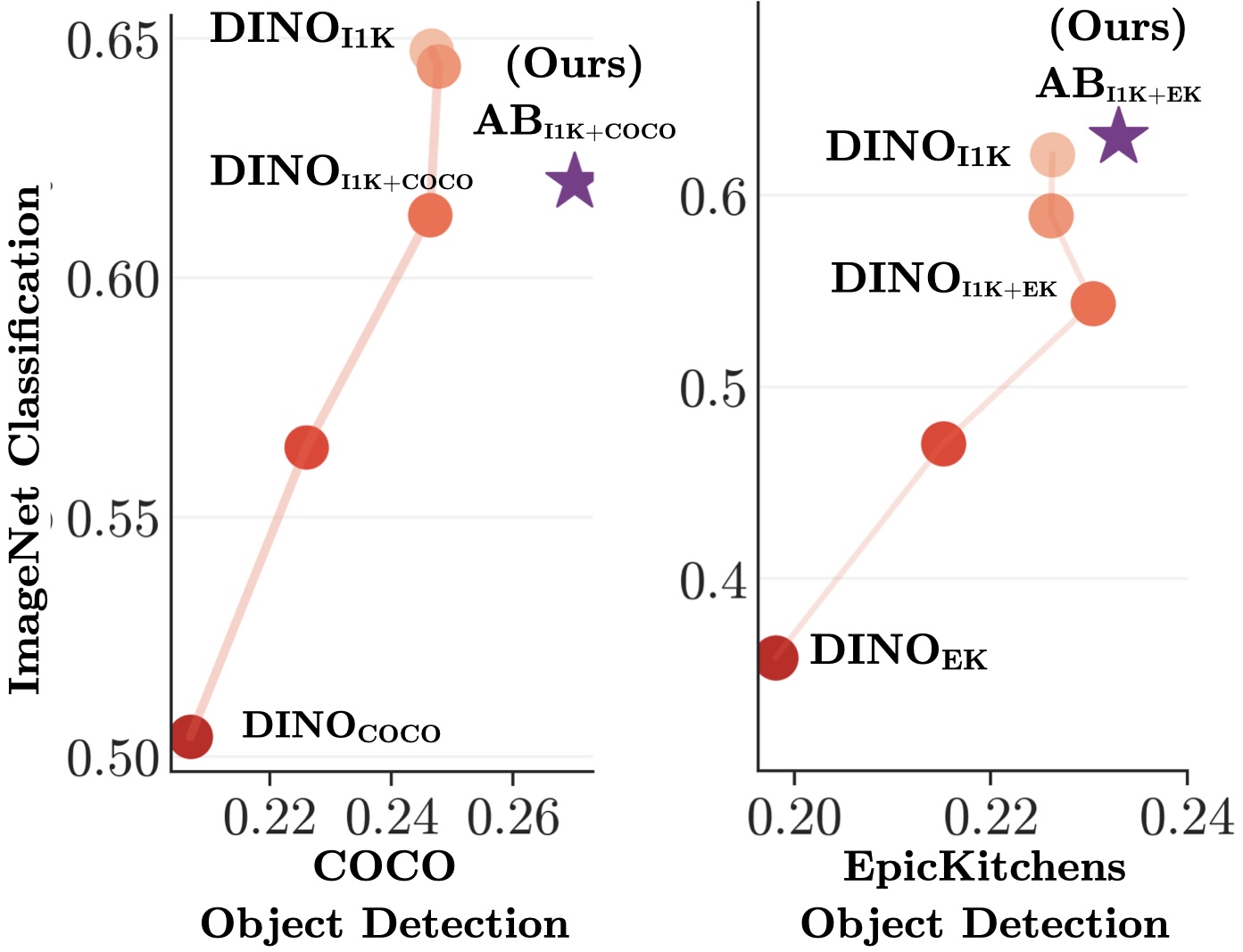}
    \vspace{-1em}
    
    \caption{We compare decoupled training of \abdino~on ImageNet and COCO or EpicKitchens to running DINO on a mixture for $p \in \{0, 0.25, 0.5, 0.75, 1.0\}$. \abdino~outperforms all DINO mixtures, indicating that value bootstrapping learns more useful features than crop-invariance on any combination of the two datasets.}
    \label{fig:mixture}
\end{wrapfigure}
These datasets are a poor fit for the inductive bias underlying consistency methods, containing many (small) objects, and crop-consistency methods like DINO and SimCLR underperform relative generic criteria like MAE pixel prediction(Table \ref{table:coco_and_k}). Annotation bootstrapping improves over only running the base loss, but remains slightly worse than MAE, indicating that while bootstrapping can improve features, it cannot significantly improve upon a base reward that does not capture semantic details well.

We test the ability of annotation bootstrapping to decouple the annotation and bootstrapping data distributions, since in theory the former loss may be optimized with a curated dataset to specify semantics, and learning only by bootstrapping on our target unlabeled images. In Table \ref{table:coco_and_k}, below the line, we find that this decoupled approach leads to significantly better performance for in-domain tasks like object recognition, localization, and action recognition. We analyze this more carefully by sweeping a base DINO algorithm with 5 different data mixture ratios between Imagenet and \{ Coco, EpicKitchens \}. Our results in Figure \ref{fig:mixture}, indicate that \abdino~learns representations better than the frontier generated by DINO.

\input{tables/coco}

\input{tables/ek}
\paragraph{Pre-training with a weakly-supervised base loss.}

We next evaluate annotation bootstrapping in the weakly labeled setting, when the annotations are tokenized text captions. Recall that in this setting, our approach learns by associating text from images using a base CLIP loss, and bootstrapping by estimating value predictions about image-text relationships of other crops of an unlabeled image.

\begin{figure}
    \centering
    \includegraphics[width=\linewidth]{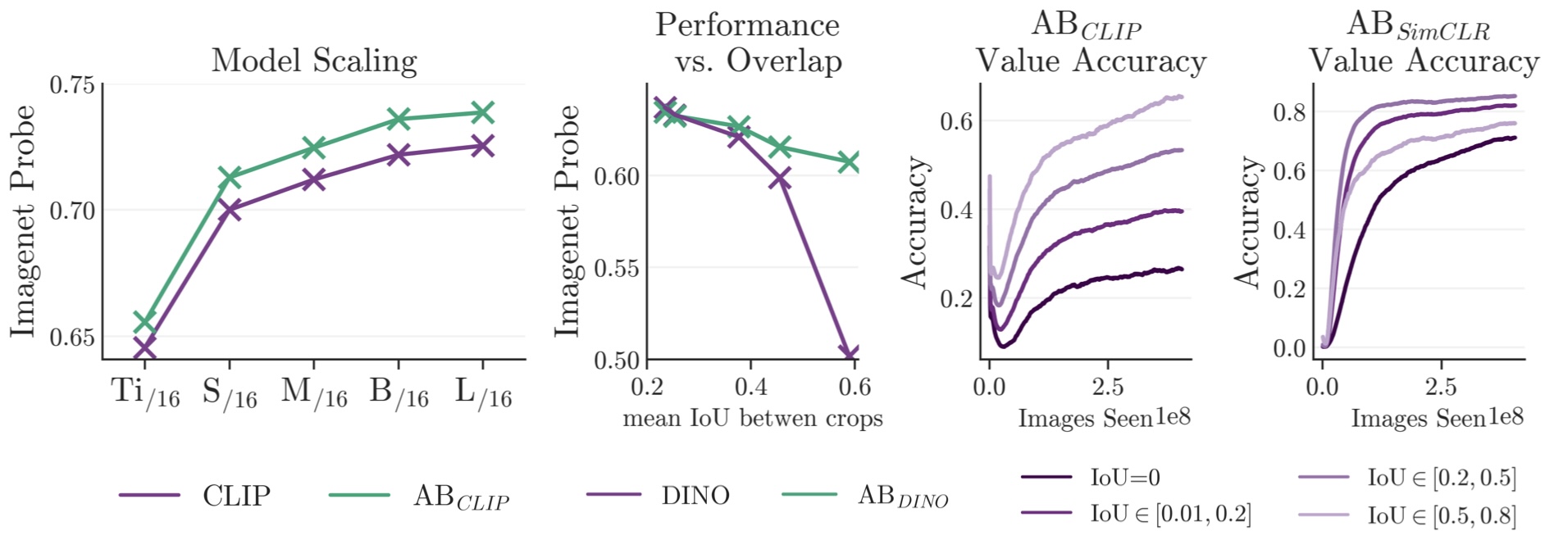}
    \vspace{-2em}
    \caption{(left) \abclip~improves over CLIP across a range of model scales (middle) Controlling the difficulty of the bootstrapping prediction problem, performance degrades as the overlap source and target crop grows, but slower than vanilla DINO. (right). Accuracy of annotation decoder in training; the model quickly predicts annotations for nearby crops, and slowly learns about more distant ones.}
    \vspace{-1em}
    \label{fig:clip-crops-training}
\end{figure}

On CC12M (Table \ref{table:cc12m}, bottom), we see that weakly supervised methods across the board outperform their unsupervised equivalents; this matches empirical evidence that contrastive language-text methods are more capable of training on lower-quality image data. As discussed by \citet{naeem2023silc}, we find that combining CLIP with a self-supervised objective like DINO (SiLC) or SimCLR (SLIP) primarily improves fine-grained reasoning on the ClevR benchmark tasks, with only marginal improvement on downstream classification tasks. In contrast, annotation bootstrapping obtains much stronger performance relative to these other approaches on classification and segmentation metrics we evaluated, in particular improving by $\geq 4\%$ on downstream ImageNet probing performance over the base CLIP representations. Plotting the gradient alignment between the CLIP and self-supervised losses in Appendix \ref{app:analysis_and_learning_dynamics}, we see that the \abclip~objective has higher correlated than the SLIP and SILC losses, implying that our self-supervision is better aligned to semantic features relevant to CLIP. In Figure \ref{fig:clip-crops-training} (left), we investigate model scaling using up to $\approx 100\times$ more compute; across these model scales, \abclip consistently improves over CLIP training.

We further compare different weakly-supervised methods for pre-training on COCO in Table \ref{table:coco_clip}, a dataset where we found crop-consistency methods to struggle. We source text descriptions of these images from two annotation sources: captions \citep{DBLP:conf/cvpr/KarpathyL15} and bounding box descriptions \citep{Lin2014MicrosoftCC}, both directly present in the COCO dataset. 

\input{tables/coco_clip}

In this setting, only \abclip~ improves over CLIP, while both SLIP and SiLC counterintuitively \textit{decrease} in performance. Our findings support the hypothesis of \citet{Weers_2023_CVPR}, that invariance-based objectives are not necessarily additive upon weakly supervised learning, but instead move the model towards an invariant solution. When crop-consistency matches the inductive biases of the data, adding self-supervision leads to improved performance, but otherwise may degrade. In contrast, annotation bootstrapping seems to improve performance over the base CLIP loss, even when inductive biases like consistency do not fit the unlabeled data.

\subsection{Analysis} 	

We now more carefully investigate the learning dynamics of \ab~and the effect of design decisions in the method. These ablatory experiments are run using a budget of 400M views. See Appendix \ref{app:analysis_and_learning_dynamics} for visualizations of the learned value functions and prediction targets through training.

\textit{How well is the value bootstrapping objective optimized through training?}  In Figure \ref{fig:mixture} (right), we plot the prediction accuracy for the bootstrapping objective throughout the course of training, clustering by how far the target prediction box is in terms of IoU. Notice that prediction errors increase initially in training as the annotation head is first learned, but decreases uniformly through training. We note that the prediction problem is more challenging for \abclip than for \abdino, reflecting the fact that the base DINO objective is jointly mak the predictive distribution more similar across different crops, while CLIP learns a fixed and grounded annotation space.

\textit{How does the choice of crops affect the quality of bootstrapping?} We next investigate how the choice of bounding boxes affects the performance of the algorithm, by sampling source and target bounding boxes that are closer (or further) apart while keeping the marginal distribution over bounding boxes fixed. In Figure \ref{fig:mixture}, we see that performance increases steadily as the average IoU between the source and target distributions is decreased, but also that \abdino is far more robust at learning from closely-related crops than DINO. Combined with Figure \ref{fig:mixture}, these results indicate that training on more distant examples offers the most useful learning signal. 
\input{tables/ablations}

\textit{What components of annotation bootstrapping most affect downstream performance?}

We ablate different components of the method in Table \ref{table:ablations}. As with other bootstrapping and self-distillation methods, we find that removing the EMA network nullifies all performance gains from the bootstrapping objective. Similarly, removing the base loss, which grounds annotation distributions in a semantically meaningful space, significantly degrades performance. We also perform an ablation replacing the bounding box description tokens with empty mask tokens, thereby forcing the model to predict the \textit{average} annotation distribution across different crops. Doing so turns the bootstrapping objective from one of equivariance to invariance, since all crops are trained to match the same average distribution -- this again degrades performance.

\section{Discussion}

Our paper explored using RL techniques to pre-train visual representations on unlabeled data. Just as value functions predict rewards over extended temporal horizons, our method similarly learns by predicting annotations associated with ``distant'' sub-crops of an image through temporal-difference learning. Annotation bootstrapping partitions learning into the specification of image semantics (the reward loss) and bootstrapping (the value loss), allowing us to learn useful features using curated or labeled datasets, while still training on unlabeled images that lack the same supervision or inductive biases as the curated data. Across several datasets, annotation bootstrapping learned useful representations beyond those from objectives like pixel prediction, CLIP, or crop invariance. 

While the bootstrapping objective reduces the dependency on inductive biases compared to invariance-based approaches, the choice of crops remains an important factor in the quality of learned representations. Annotation bootstrapping also is not standalone: rather, it accentuates a base reward loss like CLIP or DINO to learn better features than the base loss alone. There is a lot that remains to be answered: how RL objectives behave for larger models and data scales, whether other RL algorithms may be more competent feature learners, or even how non-contrastive versions of bootstrapping may enable weak supervision beyond captions or images. Annotation bootstrapping explores one approach to learn from unlabeled images in a more targeted and semantically meaningful manner. As our largest models now saturate weakly-labeled data scraped from the internet, it seems inevitable that the largely untapped banks of unlabeled image data becomes the next frontier for visual pre-training.

\subsection*{Acknowledgements}

The authors thank Katie Kang, Colin Li, Oier Mees, Sudeep Dasari, Manan Tomar, Philippe Hansen-Estruch, and the members
of RAIL for discussions and helpful feedback. The
research was supported by the TPU Research Cloud and the Office of Naval Research (ONR N00014-22-1-2773).

\bibliographystyle{abbrvnat}
\bibliography{iclr2025_conference}

\clearpage
\appendix

\input{appendix}

\end{document}

%% file: math_commands.tex
\usepackage{amsmath,amsfonts,bm}

\def\eqref#1{equation~\ref{#1}}

\def\1{\bm{1}}

\DeclareMathAlphabet{\mathsfit}{\encodingdefault}{\sfdefault}{m}{sl}
\SetMathAlphabet{\mathsfit}{bold}{\encodingdefault}{\sfdefault}{bx}{n}

\def\gB{{\mathcal{B}}}

\def\gL{{\mathcal{L}}}
\def\gM{{\mathcal{M}}}

\newcommand{\E}{\mathbb{E}}

\let\ab\allowbreak

%% file: my_algorithm.tex
\begin{algorithm}[t]
\caption{Annotation Bootstrapping (visualized in Figure \ref{fig:bootstrapping}, pseudocode in Appendix \ref{app:algo}.)} \label{alg:mainalg}
\begin{algorithmic}
\Require Batch of unlabeled images $\gB_{u} = \{x_k\}$, batch of reward annotations $\gB_a = \{(x_i, \ell_i)\}$
\State
\State Compute reward loss to contrastively estimate annotation distribution $p(\ell|x)$, e.g. using InfoNCE.
\begin{equation*}
    \gL_{reward} = -\sum_{i} \log \frac{\exp(f(x_i, \ell_i))}{\sum_j \exp(f(x_i, \ell_j)} + \log \frac{\exp(f(x_i, \ell_i))}{\sum_j \exp(f(x_j, \ell_i)}
\end{equation*}
\State Sample $n$ random crops from unlabeled images $\{(x_k^1, \bb_1), \dots, (x_k^n, \bb_n)\}$
\State For each image crop $j \in \{1, \dots, n\}$ and annotation $\ell \in \gB_a$, compute target value estimate:
\begin{equation*}
     \hat{Q}_k^j(\ell) = (1-\gamma)\operatorname{softmax}(f(x_k^j, \ell)) + \gamma \max_{a \in \bb_{j \to *}}Q^\ema(x_k^j, a, \ell)
\end{equation*}
\State For each image crop $i \in \{1, \dots, n\}$, optimize Equation \ref{eq:bootstrapping} with $(x=x_k^i, a=\bb_{i \to j}, x'=x_k^j)$ 
\begin{equation*}
    \gL_{\ab} = \operatorname{BinaryCE}(\hat{Q}_k^j(\ell), Q(x_k^i, \bb_{i \to j}, \ell) )
\end{equation*}
\end{algorithmic}
\end{algorithm}

%% file: tables/i1k_probe.tex
\begin{wraptable}{r}{0.6\textwidth}
\small
\begin{tabularx}{\linewidth}{ccYYYY}
\toprule
& \Th{Method} & \Th{Linear} & \Th{MAP} & \Th{Decoder} & \Th{Finetune}\\ \midrule
& MAE & 47.8 & 63.4 & 66.5 & 79.0 \\
 & I-JEPA & 58.5 & 61.5 & 64.5 & 76.7 \\
ImageNet & SimCLR & 66.2 & 69.7 & 71.4 & 76.8 \\
\scriptsize{No} & \absimclr \scriptsize{(Ours)} & 64.5$_{-1.7}$ & 70.7$_{+1.1}$ & 72.2$_{+0.8}$ & 77.0$_{+0.2}$ \\
\scriptsize{Labels} & DINO & \textbf{68.9} & 70.9 & 72.7 & \textbf{79.7} \\
 & \abdino \scriptsize{(Ours)} & 67.2$_{-1.7}$ & \textbf{73.1$_{+2.2}$} & \textbf{74.8$_{+2.1}$} & \textbf{80.3$_{+0.6}$} \\\bottomrule
\end{tabularx}
\caption{Bootstrapping a self-supervised loss learns better representations than training with the base loss alone on unlabeled ImageNet images, especially beyond linear probes.}
\label{table:i1k_probe}

\vspace{-1em}
\end{wraptable}

%% file: tables/cc12m.tex
\begin{table*}
\centering
\small
\setlength{\tabcolsep}{3pt}
\caption{Bootstrapping annotations improves over several weakly supervised and self-supervised base losses (\abclip~over CLIP, \absimclr~over SimCLR, \abdino~over DINO) on CC-12M \citep{Changpinyo2021Conceptual1P}, a weakly-curated web crawl dataset with $8.7$ million images. The gap is greatest for \abclip, where it significantly outperforms other approaches combining self-supervision with captions.  \textbf{*Avg. Cls} averages classification accuracy over the benchmarks in \citet{Beyer2023ASO}: Food101, Oxford IIIT Pets, Resics45, and Sun397.}
\label{table:cc12m}
\begin{tabularx}{\linewidth}{YXlllll}
\toprule
\Th{Pretrain Dataset} & \Th{Method} & \Th{ImageNet} & \Th{Avg Cls*} & \Th{Clevr$_\text{/Depth}$} & \Th{Clevr$_{\text{/Count}}$} \\ \midrule
 & MAE & 61.3 & 75.4 & \textbf{82.8} & \textbf{90.4} \\
 & I-JEPA & 60.0 & 76.0 & 80.1 & \textbf{90.0} \\
CC12M & SimCLR & 67.3 & 79.0 & 76.5 & 89.4 \\
(no captions) & \absimclr (Ours) & 68.0$_{+0.7}$ & 79.5$_{+0.4}$ & 79.5$_{+3.0}$ & \textbf{89.6$_{+0.2}$} \\
 & DINO & 68.9 & 80.9 & 79.3 & 87.6 \\
 & \abdino (Ours) & \textbf{70.6$_{+1.8}$} & \textbf{82.2$_{+1.3}$} & 80.4$_{+1.1}$ & \textbf{89.9$_{+2.4}$} \\
                                      \midrule
& CLIP & 69.5 & 82.8 & 70.0 & 84.4 \\
 & CLIP \tiny{+Aug} \citep{fini2023improved} & 72.6 & 85.0 & 72.7 & 87.0 \\
 & SLIP \tiny{+SimCLR} \citep{Mu2021SLIPSM} & 72.0 & 84.3 & 72.4 & 87.2 \\
 & SiLC \tiny{+DINO} \citep{naeem2023silc} & 72.8 & 85.0 & 74.4 & 88.2 \\
 & \abclip (Ours) & \textbf{74.1$_{+4.6}$} & \textbf{85.6$_{+2.8}$} & \textbf{78.1$_{+8.1}$} & \textbf{91.9$_{+7.4}$} \\
      \bottomrule
\end{tabularx}
\end{table*}

%% file: tables/coco.tex
\begin{table}
\centering
\scriptsize
\setlength{\tabcolsep}{3pt}
\caption{{Since COCO \citep{Lin2014MicrosoftCC} and EpicKitchens \citep{Damen2020TheED} are not object-centric datasets like ImageNet, invariance-based self-supervised methods like SimCLR and DINO degrade. \absimclr~and \abdino~can alleviate these deficiencies. Since reward and value learning is decoupled in \ab, we improve feature learning by learning rewards from ImageNet images and training the value function to bootstrap this reward signal on COCO and EpicKitchens.}}
\label{table:coco_and_k}
\begin{minipage}[t]{0.46\linewidth}
\centering
\fontsize{7}{7}
\begin{tabularx}{\linewidth}{lllll}
\toprule
& {\scriptsize{\textsc{Method}}} & {\scriptsize \makecell{\textsc{Imagenet}\\ \textsc{Cls.}}} & {\scriptsize \makecell{\textsc{COCO}\\\textsc{Object}\\\textsc{Detect}}}
 & {\scriptsize \makecell{\textsc{COCO}\\\textsc{Object}\\\textsc{Cls.}}} \\ \midrule
 & MAE & 58.1 & \textbf{29.8} & \textbf{73.8} \\
 & I-JEPA & 43.0 & 21.0 & 62.5 \\
 & SimCLR & 55.0 & 20.5 & 69.7 \\
COCO & \absimclr (Ours) & \textbf{60.7$_{+5.7}$} & 26.4$_{+5.8}$ & \textbf{74.0$_{+4.3}$} \\
 & DINO & 51.0 & 21.0 & 66.9 \\
 & \abdino (Ours) & 58.6$_{+7.6}$ & 26.1$_{+5.1}$ & 72.8$_{+5.9}$ \\
 \midrule
COCO + & \absimclr & \textbf{68.3} & \textbf{31.0} & \textbf{79.4}\\
ImageNet  & \abdino & 65.2 & \textbf{31.0} & \textbf{78.8} \\
\bottomrule
\end{tabularx}
\end{minipage}
\hfill
\begin{minipage}[t]{0.53\linewidth}
\centering
\fontsize{7}{7}
\begin{tabularx}{\linewidth}{llcccc}
\toprule
& \Th{\scriptsize Method} & \Th{\scriptsize\makecell{Imagenet \\ Cls.}} & \Th{\scriptsize \makecell{EK\\Action\\Recog.}} & \Th{\scriptsize \makecell{EK\\Object\\Detect}} & \Th{\scriptsize \makecell{EK\\Object\\Cls.}} \\ \midrule
 & MAE & 38.1 & 19.6 & \textbf{37.8} & \textbf{44.3} \\
 & I-JEPA & 38.7 & 18.5 & 28.0 & 39.5 \\
Epic & SimCLR & 44.5 & \textbf{21.6} & 28.7 & 40.2 \\
Kitchens & \absimclr (Ours) & \textbf{46.7$_{+2.2}$} & 20.9$_{-0.6}$ & 32.0$_{+3.3}$ & 43.2$_{+3.0}$ \\
 & DINO & 42.3 & 20.2 & 26.5 & 38.6 \\
 & \abdino (Ours) & 45.5$_{+3.2}$ & \textbf{21.6$_{+1.4}$} & 32.6$_{+6.1}$ & 43.2$_{+4.6}$ \\
\midrule
EK + & \absimclr & \textbf{68.4} & \textbf{23.5} & \textbf{39.2} & {47.3}\\
ImageNet & \abdino & 62.7 & {22.8} & {36.1} & \textbf{47.6}\\
\bottomrule
\end{tabularx}
\end{minipage}
\end{table}

%% file: tables/coco_clip.tex
\begin{wraptable}[14]{l}{0.5\linewidth}
\centering
\setlength{\tabcolsep}{3pt}
\caption{Combining CLIP with SimCLR or DINO on COCO degrades performance; only \abclip~learns better features than vanilla CLIP.}
\label{table:coco_clip}
\small
\begin{tabular}{clll}
\toprule
\Th{\scriptsize{Annotation Type }} & \Th{\scriptsize Method} & \Th{\scriptsize {Detection}} & \Th{\scriptsize {Object Cls}} \\ \midrule
 & CLIP & 26.1 & 71.2 \\
COCO & SLIP \tiny{+SimCLR} & 25.6 & 72.9 \\
Captions & SiLC \tiny{+DINO} & 25.6 & 73.2 \\
 & \abclip (Ours) & \textbf{27.8$_{+1.7}$} & \textbf{74.7$_{+3.5}$} \\
 \midrule
 & CLIP & 25.7 & 70.0 \\
Bounding & SLIP \tiny{+SimCLR} & 25.2 & 72.3 \\
Boxes & SiLC \tiny{+DINO} & 26.0 & 73.8 \\
 & \abclip (Ours) & \textbf{28.8$_{+3.1}$} & \textbf{76.8$_{+6.8}$} \\
\bottomrule
\end{tabular}
\end{wraptable}

%% file: tables/ablations.tex
\begin{wraptable}[11]{r}{0.5\linewidth}
\centering
\small
\begin{tabularx}{\linewidth}{YY}
\toprule
\Th{Ablation} & \Th{\makecell{Imagenet\\Performance}} \\ \midrule
\absimclr & 62.9~~~\\
Adding augmentations & 62.5 $_{-0.4}$\\
Removing action tokens & 60.5 $_{-2.4}$\\
No propagation loss    & 59.4 $_{-3.5}$\\
No target network & 59.4 $_{-3.5}$\\
No annotation loss & 39.0 $_{-23.9}$\\
\bottomrule
\end{tabularx}
\caption{Ablations of \absimclr on CC12M.}
\label{table:ablations}
\end{wraptable}

%% file: appendix.tex
\section{Theoretical Foundations}
\label{app:theory}

\subsection{The RL Environment}
\label{app:rl}

The agent acts in an MDP $\gM$, where agents take actions that stochastically ``transform'' image views $x' \sim P(x' | x, a)$ and, for any given task $\ell$, optimizes the likelihood of seeing the annotation $p(\ell|x)$ under views sampled from the time-discounted visitation distribution of the agent,

\begin{equation}
    \max_{\pi} p(\ell | x_+) = \E_{x_t \sim d^\pi}[p(\ell | x_t)] = \E\left[\sum_{t=1}^\infty \gamma^t \underbrace{(1-\gamma)p(\ell | x_t)}_{\text{reward}} ~~|~~ a_t \sim \pi(\cdot | x_{t}, \ell),  x_{t+1} \sim P(\cdot | x_t, a_t)\right].
\end{equation}

This formulation is a common way of defining MDPs for goal-achieving tasks \citep{blier2021learning, rudner2021outcome, Eysenbach2022ContrastiveLA}, and is also related to a rich literature in active perception \citep{bajcsy1988active, aloimonos1988active,  papanikolopoulos1991vision, rivlin2000control, jayaraman2018learning}. Optimal behavior in this environment is defined by two distributions: the reward $p(\ell | x)$ and the transition dynamics $P(x' | x, a)$. The former is defined through the ``base'' loss function (e.g. CLIP, SimCLR, or DINO), and the latter is induced implicitly through the choice of random cropping mechanism. Formally, sampling two ``views'' from a parameterized image augmentation, $T: I, c \mapsto x$ induces a distribution $P(c_1, x_1, c_2, x_2)$, which defines the transition dynamics: $P(x_2 | x_1, \underbrace{(c_1, c_2)}_{\text{action}})$. We use random crop transformations, $(c_1, c_2)$, which correspond to bounding boxes $(\bb_1, \bb_2)$, and so we represent actions as $\bb_{1 \to 2}$. In principle, however, this can be done with any sampling strategy.

Q-learning methods \citep{watkins1992, Mnih2013PlayingAW} learn the optimal policy by iteratively applying a Bellman operator $(\mathcal{T}Q)(x, a) = \E_{x'}[r(x') + \gamma \max_{a'} Q(x', a')]$; in the tabular setting, this operator is contractive with a unique fixed point at $Q^*$, the value function of the optimal policy. Q-learning methods are particularly appealing because they are \textbf{off-policy}, enabling the use of arbitrary transitions $(x, a, x')$ for learning. In practice, this allows us to train from samples from our fixed random crop strategy,  without needing to sample from our ``policy'' (which would require significantly greater infrastructure to set up). With off-policy methods, we are able to effectively keep the data pipeline exactly like other self-supervised methods, with no extra tweaks needed.

Our objective is a near-vanilla implementation of deep Q-learning \citep{Mnih2013PlayingAW}, learning to predict outputs generated by the Bellman operator applied to a lagging EMA model, using transition samples $(x, a, x')$ generated by the random cropping process and tasks $\ell$ sampled from the marginal distribution over annotations:

\begin{equation}
    \min \E_{\substack{I \sim D_u \\ \{(x_i, \bb_i)\}_{i=1}^n \sim \operatorname{RRC}(I) \\ \ell \sim p(\ell)}} \left[\sum_{i,j} D(Q(x_i, \bb_{i \to j}, \ell), (1-\gamma)p(\ell| x_j) + \gamma \max_{k} Q^\ema(x_j, \bb_{j \to k}, \ell))\right].
\end{equation}

In our setting, we cannot analytically compute the maximum, so we follow the technique of \cite{emaq} and estimate over $n$ sampled actions; this approximation has connections to maximum-entropy RL and other soft variants \citep{ziebart2008maximum, kostrikov2021offline}.

\subsection{Annotation Bootstrapping Details}
\label{app:algo}
\label{app:connections}

In this section, we detail the model and training procedure used for annotation bootstrapping (See loss code in Algorithm \ref{alg:actual_loss}). We emphasize that even though our approach optimizes a reinforcement learning objective, the actual model architecture and training code greatly resemble other self-supervised and semi-supervised learning methods. When relevant, we highlight similarities and differences in technical details with other self-supervised approaches.

\begin{figure}
    \centering
    \includegraphics[width=\linewidth]{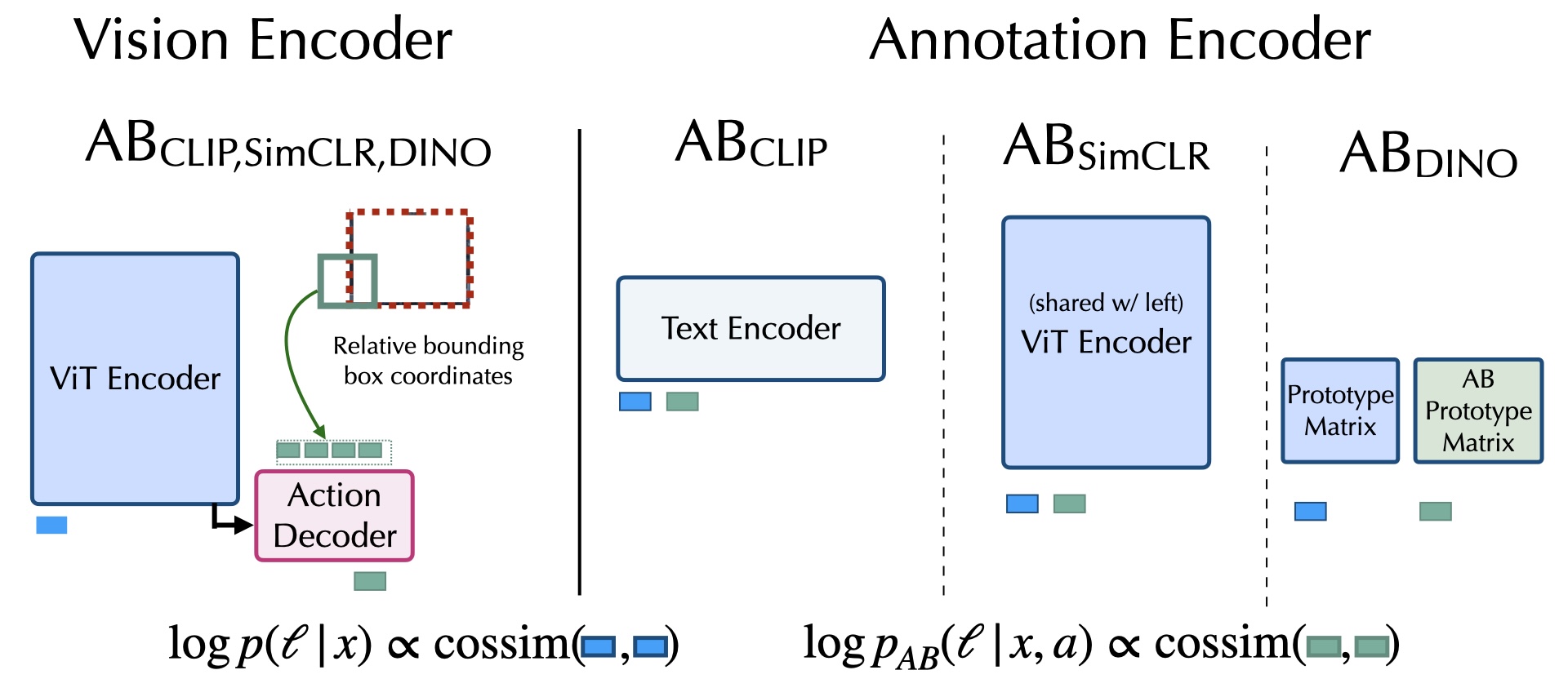}
    \caption{Visualizations of Annotation Bootstrapping for different base learning algorithms, \abclip, \absimclr, \abdino. The model architectures are near-identical on the visual side: a ViT vision encoder, with a head for the base loss, and a Decoder transformer to predict the annotations associated with other bounding boxes. What differs between the implementations is how annotations are embedded. In CLIP, they are embedded by a separate text encoder; in SimCLR, they are embedded by the same vision backbone; in DINO, annotations come from a weight matrix of prototypes. All methods train with the recipe in Algorithm \ref{alg:actual_loss}, inserting the corresponding reward loss. }
    \label{fig:enter-label}
\end{figure}

\textbf{Model:} On the visual side, the model (visualized in Figure \ref{fig:enter-label}) consists of a vision encoder followed by a four-layer $S$-sized transformer decoder, that attends to a set of action tokens and cross-attends to the final outputs of the vision transformer. Actions transforming a view $i$ to $j$ are parameterized as a sequence of four discretized tokens $a = \bb_{i \to j} = (y_{\min}, x_{\min}, y_{\max}, x_{\max})$ that describes the relative coordinates of the view $j$ with respect to the view $i$. In using a fixed number of tokens to represent each action, we are able to easily run the decoder for several actions on each image with minimal overhead relative to the rest of the network. This transformer decoder is similar to those used for  I-JEPA \citep{Assran2023SelfSupervisedLF} and CrossMAE \citep{fu2024rethinking}, except for the use of a fixed number of tokens to describe the action (c.f. target view in I-JEPA, CrossMAE) instead of using one token per patch in the target view.

\begin{wrapfigure}{r}{0.6\linewidth}
    \centering    \includegraphics[width=\linewidth]{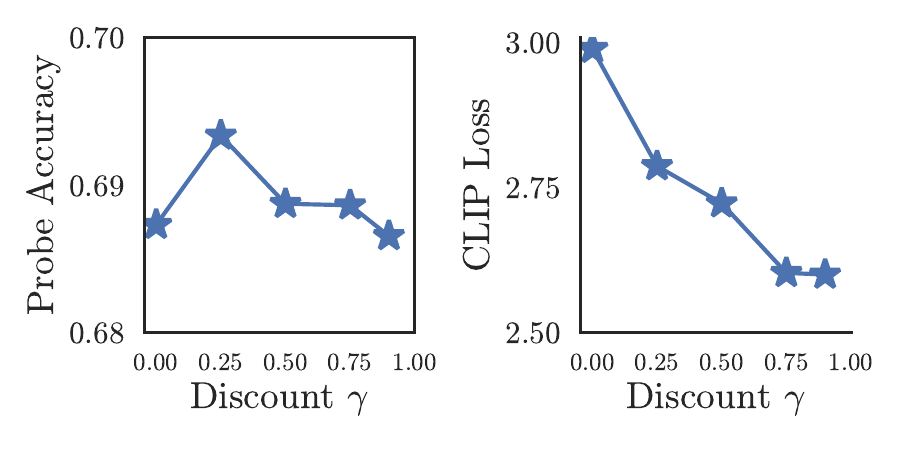}
    \caption{Effect of discount factor $\gamma$ on \abclip; (left) Imagenet probing accuracy, (right) CLIP reward loss. We use $\gamma=0.5$ for our main experiments as a sensible default.}
    \label{fig:gamma}
\end{wrapfigure}

\textbf{Objective:} The loss function used in annotation bootstrapping is described in Algorithm \ref{alg:actual_loss}. There is one primary hyperparameter relative to other methods, the discount factor $\gamma$. In Figure \ref{fig:gamma}, we sweep across different settings of $\gamma$ for \abclip -- in general, we find that intermediate choices of $\gamma$ offer a tradeoff between probing accuracy and zero-shot CLIP performance. We choose to use $\gamma=0.5$ for our experiments as a sensible default, equally balancing between the current reward $(1-\gamma)p(\ell|s)$ and the future predictions from the next-state value network $\gamma Q(s', a', \ell)$. We note that the Bellman operator $r + \gamma \max (\cdot)$ provides a natural asymmetry to the self-distillation process; serving a similar purpose to the teacher logit sharpening in DINO and the asymmetric prediction heads for BYOL.

\textbf{Training:} We use a lagging EMA model to generate target predictions, similar to other self-distillation \citep{Caron2021EmergingPI, Grill2020BootstrapYO} and RL \citep{Mnih2013PlayingAW} approaches. In the fully self-supervised setting, we follow the recommendations of \citet{Caron2021EmergingPI}, with an EMA parameter $\tau$ that decays from $0.004$ to $0$ following a cosine schedule through training; in the setting with weak supervision, we follow \citet{naeem2023silc}, updating the EMA model instead with a constant schedule of $\tau=0.004$.

\begin{algorithm*}
\caption{General Annotation Bootstrapping Pseudocode}\label{alg:actual_loss}
\begin{minted}[fontsize=\footnotesize]{python}

def loss(reward_batch, value_batch, model, ema_model):
  r_logits = model(reward_batch['image'], reward_batch['text'])
  # The annotation loss (CLIP here) associates images and annotations
  # Replace with SimCLR loss or DINO loss for the appropriate variants
  reward_loss = CrossEntropy(r_logits, eye(B_a)) + CrossEntropy(r_logits.T, eye(B_a))

  # The bootstrapping loss uses view1 to predict annotations associated with view2
  views, bboxes = RandomResizedCrop(value_batch['image']) x n
  # views is B x N x (H x W x c)
  actions = relative_bbox(bboxes) # B x N x N x 4
  # actions[:, j, k] is the action that transforms view j into view k
  ema_r_logits, ema_q_logits = ema_model(views, reward_batch['text'], actions=actions)
  # ema_r_logits: B x N x L, ema_q_logits: B x N x N x L
  # ema_q_logits[:, j, k, l] = estimated Q of taking action "j->k"
  #                            from view j for task annotation `l`
    target_p = (1-gamma) * softmax(ema_r_logits, axis=-1) + 
                gamma * sigmoid(ema_q_logits.max(axis=2))
  _, q_logits = model(views, reward_batch['text'], actions=actions)
  value_loss = BCE(q_logits, target_p[:, None, :, :])
  
  return reward_loss + value_loss
\end{minted}
\end{algorithm*}

\section{Training Details}
\label{app:train}
\textbf{Models.} We implement our models and baselines in JAX, using the \texttt{bigvision} repository \citep{big_vision} implementation of all transformer components, such as the vision encoder, the text encoder for CLIP, and the annotation decoder that predicts the ``value'' of future annotations from encoded image tokens and bounding box action tokens. We were unable to replicate the results from I-JEPA in our internal codebase, so we train this baseline directly using the publicly available code.  In Table \ref{table:hyperparameters}, we provide the hyperparameters for all evaluated methods; we obtained hyperparameters from the official code-bases whenever possible; for CLIP, we adopt hyperparameters from \citet{fini2023improved}, who tune the hyperparameters of CLIP for CC-12M scale training. For the ViT-S models, training takes approximately 24 hours on a TPU v4-8 chip for 800M images seen. 

\textbf{Datasets.} We evaluate on four datasets representative of the many types of unlabelled images typically available: Imagenet \citep{Russakovsky2014ImageNetLS}, a well-curated, balanced, and image-centric benchmark heavily used by prior work; CC12M \citep{Changpinyo2021Conceptual1P}, a dataset of captioned images used for vision-language pre-training that is relatively uncurated and contains a wider range of concepts than Imagenet; COCO \citep{Lin2014MicrosoftCC} a dataset of scenes each containing many (potentially small) objects, and Epic-Kitchens \citep{Damen2020TheED}, a video dataset containing many real-world scenes in homes. Note that CC12M is a dataset of links, so links deteriorate due to rot and redirects; the version we collected \citep{beaumont-2021-img2dataset} has 8.7 million images.

\section{Evaluation}
\label{app:eval}

We use the multi-task decoder-based probe from \citet{Beyer2023ASO} for the evaluations in this paper. The probe is defined as a 4-layer transformer decoder with an autoregressive decoding pattern that attends to the final outputs of the Vision Transformer through cross-attention. We choose this architecture so that we can do all of our probing tasks, whether image recognition or bounding box prediction, or classification of the object in a bounding box using a unified framework; this also represents (albeit to a much smaller scale) how vision transformers are now used in VLM architectures. We adopt all hyperparameters for training this probe from \citet{Beyer2023ASO}. 

When pre-training on Imagenet and CC12M, we probe the model on ImageNet, the Clevr/\{Count, Distance\} tasks from \citet{Zhai2019ALS}, and then on four tasks used by \citet{Beyer2023ASO}: Food101,  Oxford IIIT Pets, Resics45, and Sun397.To test robustness to random seeds, we verified across $3$ different runs of \abclip with different initializations and different data orderings. The absolute difference in performance ($\text{max seed} - \text{min seed}$) for \textsc{Imagenet} was $0.02/100$, for \textsc{Avg CLS} was $0.4/100$, for \textsc{Clevr$_{\text{/Depth}}$} was $0.4/100$, and for \textsc{Clevr$_{\text{/Count}}$} was $0.6/100$.

When pre-training on COCO, we evaluate on small object classification (in which the model is provided the coordinates of a bounding box, and asked to predict the identity of the object within that bounding box), and the corresponding detection task (in which the model must simply identify all bounding boxes corresponding to relevant objects in a scene).

When pre-training on EpicKitchens, we probe the model also on object classification (predicting the label of an object given its bounding box) and object detection (predicting bounding boxes), which we source from the ViSOR annotation set \citep{VISOR2022}. We also probe the model's ability to predict the action a human is taking given one frame of context. This problem is not exactly solvable from one frame of context, but the relative performance differences between methods nonetheless informs the quality of the learned representations.

\input{tables/i1k_full}

\input{tables/hyperparams}

%% file: tables/i1k_full.tex
\begin{table}[H]
\label{table:i1k_full}
\centering
\scriptsize
\setlength{\tabcolsep}{3pt}
\caption{Downstream classification metrics beyond ImageNet accuracy when pre-training fully unlabelled on ImageNet. \textbf{*Avg. Cls} averages the classification accuracy over the four benchmarks in \citet{Beyer2023ASO}: Food101, Oxford IIIT Pets, Resics45, and Sun397.}
\label{tab:cc12m-results}
\begin{tabularx}{\linewidth}{XXXXXXXXX}
\toprule
\Th{Dataset} & \Th{Method} & \Th{ImageNet} & \Th{Avg Cls*} & \Th{Clevr/Depth} & \Th{Clevr/Count}  \\ \midrule
& MAE & 66.4 & 78.8 & \textbf{84.4} & \textbf{90.7} \\
 & I-JEPA & 64.5 & 79.0 & 81.0 & 88.8 \\
ImageNet & SimCLR & 71.6 & 82.2 & 76.1 & 87.1 \\
(No Labels) & \absimclr (Ours) & 72.3$_{+0.7}$ & 82.8$_{+0.6}$ & 78.6$_{+2.4}$ & 88.9$_{+1.8}$ \\
 & DINO & 72.6 & 83.2 & 79.6 & 87.4 \\
 & \abdino (Ours) & \textbf{74.6$_{+1.9}$} & \textbf{84.9$_{+1.7}$} & 80.8$_{+1.3}$ & 89.8$_{+2.4}$ \\
      \bottomrule
\end{tabularx}
\end{table}

%% file: tables/hyperparams.tex
\begin{landscape}
\begin{table}[]
\scriptsize
\caption{Hyperparameters used by all algorithms in our experiments}
\label{table:hyperparameters}
\begin{tabular}{lllllllll}
\toprule
                                                                                        & \Th{MAE}                                                                & \Th{iJEPA}         & \Th{DINO}                                                                                                                                  & \Th{SimCLR}                                                                           & \Th{CLIP}                                                              & \Th{SLIP}                                                                         & \Th{SILC}                                                                       & \Th{AB}                                                                                                               \\\midrule
\begin{tabular}[c]{@{}l@{}}Effective Batch Size\\ (= Batch Size * \# Views)\end{tabular} & 8192                                                                    & 4096               & 10240                                                                                                                                      & 8192                                                                                  & 8192                                                                   & 8192                                                                              & 9216                                                                            & 8192                                                                                                                  \\
Batch Size                                                                              & 8192                                                                    & 4096               & 1024                                                                                                                                       & 4096                                                                                  & 8192                                                                   & \begin{tabular}[c]{@{}l@{}}CLIP: 4096\\ SimCLR: 2048\end{tabular}                 & \begin{tabular}[c]{@{}l@{}}CLIP: 4096\\ DINO: 512\end{tabular}                  & \begin{tabular}[c]{@{}l@{}}Annotation batch size: base algo // 2\\ Bootstrap batch size: base algo // 8\end{tabular}  \\
Number of Views                                                                         & 1                                                                       & 1*                 & 10 (2 global, 8 local)                                                                                                                     & 2                                                                                     & 1                                                                      & 2                                                                                 & 10 (2 global, 8 local)                                                          & 4 views for bootstrap batch                                                                                           \\
Model                                                                                   & ViT-S/16                                                                & ViT-S/16           & ViT-S/16                                                                                                                                   & ViT-S/16                                                                              & \begin{tabular}[c]{@{}l@{}}ViT-S/16\\ S-size Text decoder\end{tabular} & ViT-S/16                                                                          & ViT-S/16                                                                        & \begin{tabular}[c]{@{}l@{}}Follows base loss\\ "S"-sized annotation decoder\end{tabular}                              \\
Augmentations                                                                           & \begin{tabular}[c]{@{}l@{}}RRC(0.2, 1.0),\\ HorizontalFlip\end{tabular} & RRC(0.3, 1.0)      & \begin{tabular}[c]{@{}l@{}}Global: RRC(0.4, 1.0),\\ Local: RRC(0.05, 0.04)\\ HorizontalFlip\\ ColorJitter,\\ Random GrayScale\end{tabular} & \begin{tabular}[c]{@{}l@{}}RRC(0.08, 1.0)\\ HorizontalFlip\\ ColorJitter\end{tabular} & RRC(0.5, 1.0)                                                          & \begin{tabular}[c]{@{}l@{}}Follows CLIP\\ and SimCLR\\ augmentations\end{tabular} & \begin{tabular}[c]{@{}l@{}}Follows CLIP\\ and DINO\\ augmentations\end{tabular} & \begin{tabular}[c]{@{}l@{}}For unlabeled data\\ RRC(0.05, 0.5)\\ For annotation data\\ follows base loss\end{tabular} \\
Warmup Steps                                                                            & 10,000                                                                  & 40 ImageNet Epochs & 10 ImageNet Epochs                                                                                                                         & 10 ImageNet Epochs                                                                    & 1 CC12M Epoch                                                          &                                                                                   &                                                                                 & Follows base loss                                                                                                     \\
LR                                                                                      & 2.4e-3                                                                  & 1e-3               & 1e-3                                                                                                                                       & 1e-3                                                                                  & 1e-3                                                                   &                                                                                   &                                                                                 & Follows base loss                                                                                                     \\
Weight Decay                                                                            & 0.05                                                                    & 0.04 $\to$ 0.4     & 0.04 $\to$ 0.4                                                                                                                             & 0.04 $\to$ 0.4                                                                        & 0.1                                                                    &                                                                                   &                                                                                 & Follows base loss                                                                                                     \\
Gradient Clipping                                                                       & None                                                                    & None               & 1.0                                                                                                                                        & None                                                                                  & None                                                                   &                                                                                   &                                                                                 & Follows base loss                                                                                                     \\
EMA                                                                                     & None                                                                    & 0.004 $\to$ 0      & 0.004 $\to$ 0                                                                                                                              & None                                                                                  & None                                                                   & None                                                                              & 0.004                                                                           & \begin{tabular}[c]{@{}l@{}}0.004 for \abclip\\ 0.004 $\to$ 0 for \absimclr, \abdino\end{tabular}                      \\
Additional Hyperparameters                                                              & $b_2=0.95$                                                              &                    &                                                                                                                                            &                                                                                       & $b_2=0.98$                                                             & Loss Ratio = 1.0                                                                  & Loss Ratio = 1.0                                                                & Loss Ratio = 1.0               \\\bottomrule                                                                                      
\end{tabular}
\end{table}
\end{landscape}

\section{Analysis and Learning Dynamics}
\label{app:analysis_and_learning_dynamics}

\begin{figure}[H]
    \centering
    \includegraphics[width=\linewidth]{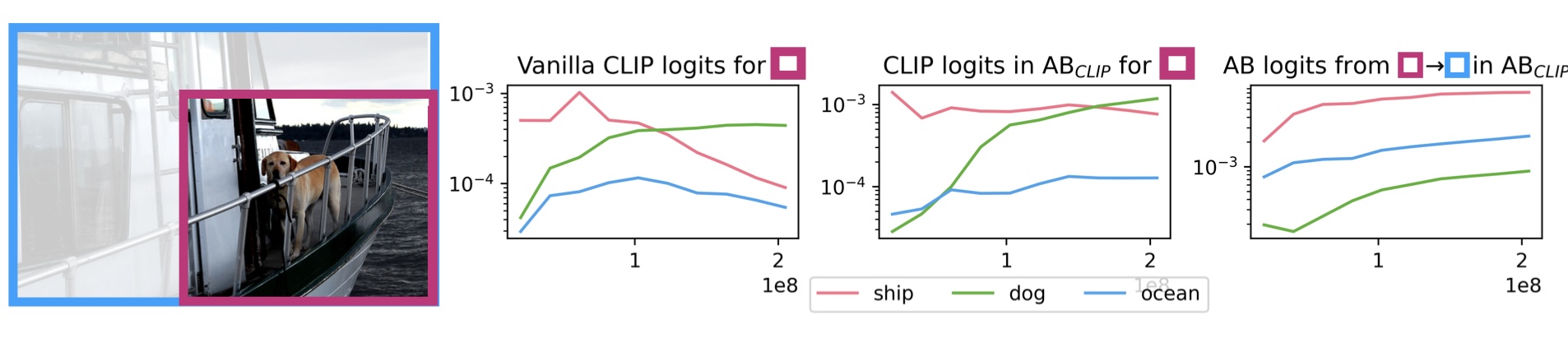}
    \caption{\textbf{Model Predictions: }An example of model predictions through training for vanilla CLIP, for \abclip, and for the value predictions of \abclip. For this particular example, vanilla CLIP recognizes early in training that the dog is on a ship, but loses this knowledge by the end of training. In contrast, \abclip does not lose this knowledge; examining the values for the ``zooming out'' action, we see that AB learns through training that the model will see a ship when the image is zoomed out.}
    \label{fig:ab_predictions}
\end{figure}

\begin{figure}[H]
    \centering
    \includegraphics[width=0.49\linewidth]{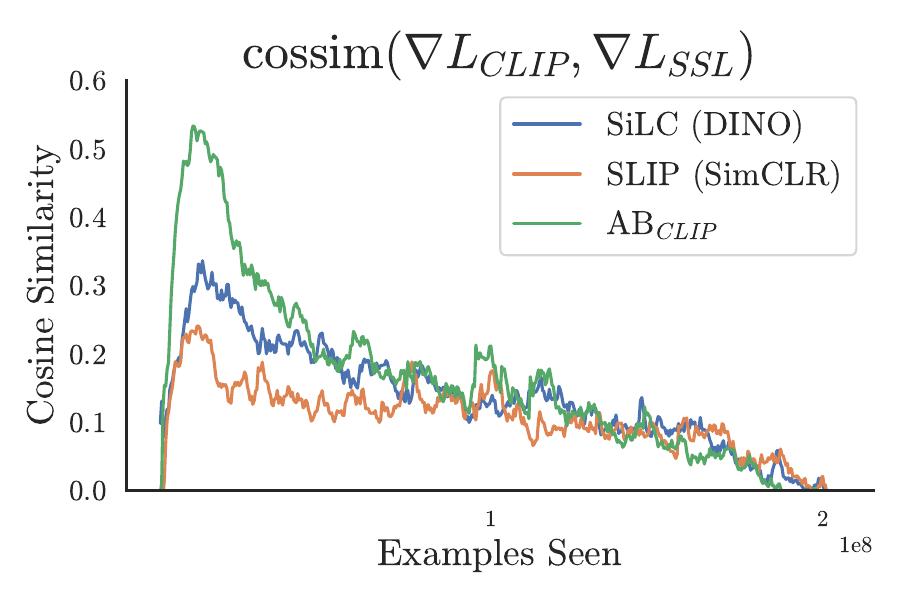}
    \includegraphics[width=0.49\linewidth]{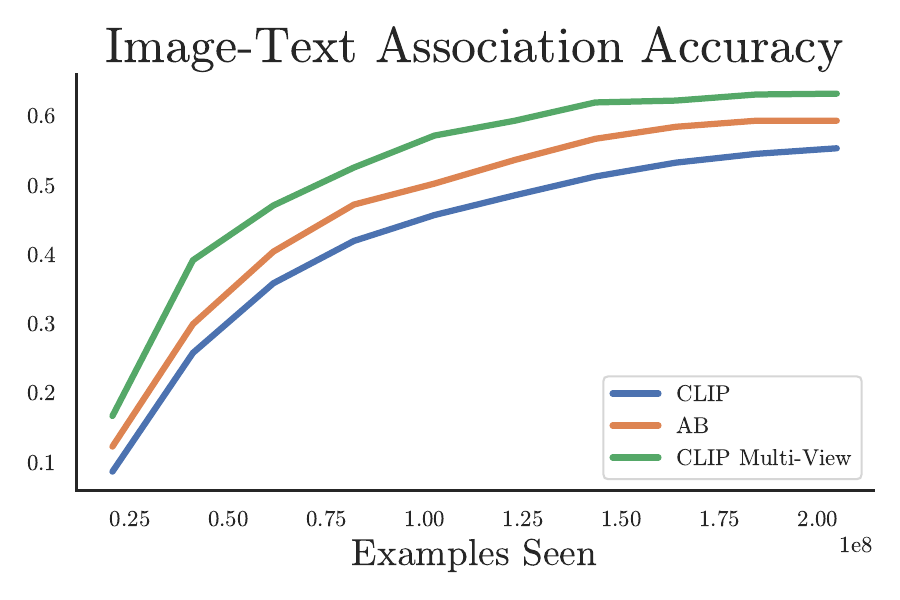}
    \caption{\textbf{Learning Dynamics:} (left) We measure the cosine similarity of ViT gradients of the self-supervised loss and the CLIP loss on held-out validation batches through training for various algorithms. While the similarity decays to zero for all methods, the value bootstrapping loss has higher cosine similarity with the CLIP objective through training, compared to SimCLR or DINO. This offers one indication that \abclip~learns features more aligned with the weakly supervised training task. (right) Through 
    \abclip~training, we measure the accuracy of ``reward'' predictions, the accuracy of the value head predictions, and the accuracy of multi-crop predictions of the reward head. The value head, which must distill the predictions of multiple different views of an image, offers a second measurement of associations between images and text beyond the vanilla CLIP loss. }
    \label{fig:learning_dynamics}
\end{figure}

\begin{figure}[H]
    \centering
    \includegraphics[width=\linewidth]{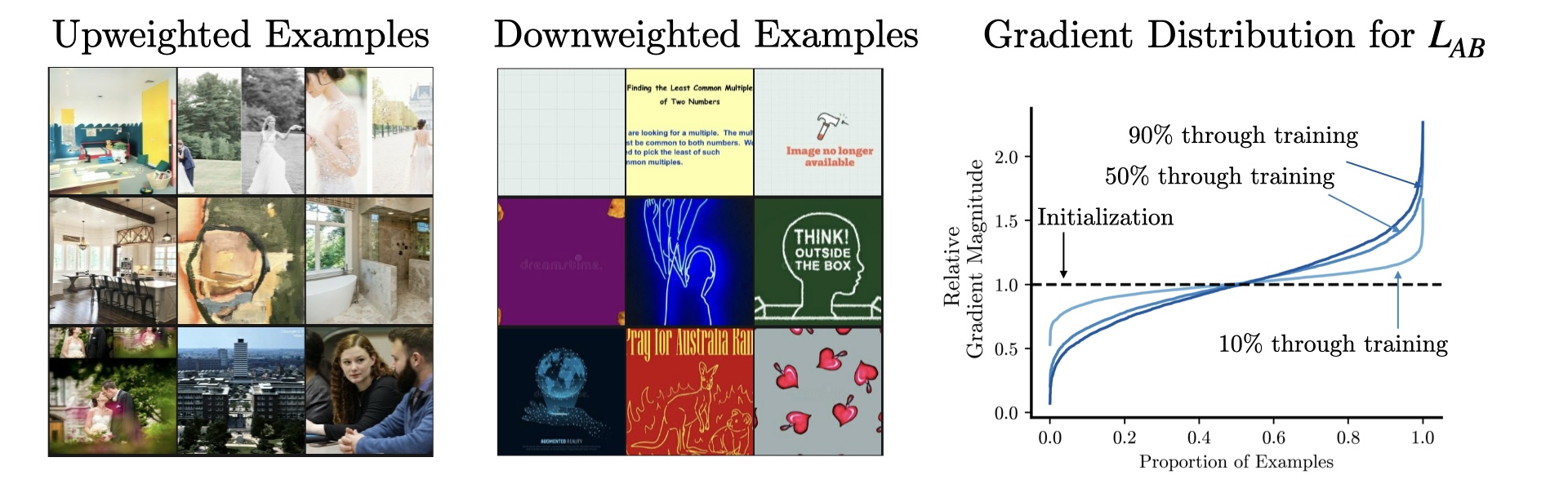}

    \caption{\textbf{Effective Curriculum:} The value loss forms an implicit curriculum over images, placing higher weight on scenes that are more visually complex, and down-weighting plain single-color images and signs.  In the plot on the right, we plot the CDF of $|\frac{\partial L_{\text{value}}}{\partial \phi}|$ at various stages of training, noting that an effective curriculum emerges by the middle of training.}
    \label{fig:effective_curriculum}
\end{figure}

%% file: neurips_2025.bbl
\begin{thebibliography}{68}
\providecommand{\natexlab}[1]{#1}
\providecommand{\url}[1]{\texttt{#1}}
\expandafter\ifx\csname urlstyle\endcsname\relax
  \providecommand{\doi}[1]{doi: #1}\else
  \providecommand{\doi}{doi: \begingroup \urlstyle{rm}\Url}\fi

\bibitem[Aloimonos et~al.(1988)Aloimonos, Weiss, and Bandyopadhyay]{aloimonos1988active}
J.~Aloimonos, I.~Weiss, and A.~Bandyopadhyay.
\newblock Active vision.
\newblock \emph{International journal of computer vision}, 1:\penalty0 333--356, 1988.

\bibitem[Assran et~al.(2023)Assran, Duval, Misra, Bojanowski, Vincent, Rabbat, LeCun, and Ballas]{Assran2023SelfSupervisedLF}
M.~Assran, Q.~Duval, I.~Misra, P.~Bojanowski, P.~Vincent, M.~G. Rabbat, Y.~LeCun, and N.~Ballas.
\newblock Self-supervised learning from images with a joint-embedding predictive architecture.
\newblock \emph{2023 IEEE/CVF Conference on Computer Vision and Pattern Recognition (CVPR)}, pages 15619--15629, 2023.

\bibitem[Bai et~al.(2024)Bai, Geng, Mangalam, Bar, Yuille, Darrell, Malik, and Efros]{bai2024sequential}
Y.~Bai, X.~Geng, K.~Mangalam, A.~Bar, A.~L. Yuille, T.~Darrell, J.~Malik, and A.~A. Efros.
\newblock Sequential modeling enables scalable learning for large vision models.
\newblock In \emph{Proceedings of the IEEE/CVF Conference on Computer Vision and Pattern Recognition}, pages 22861--22872, 2024.

\bibitem[Bajcsy(1988)]{bajcsy1988active}
R.~Bajcsy.
\newblock Active perception.
\newblock \emph{Proceedings of the IEEE}, 76\penalty0 (8):\penalty0 966--1005, 1988.

\bibitem[Bao et~al.(2021)Bao, Dong, and Wei]{Bao2021BEiTBP}
H.~Bao, L.~Dong, and F.~Wei.
\newblock Beit: Bert pre-training of image transformers.
\newblock \emph{ArXiv}, abs/2106.08254, 2021.
\newblock URL \url{https://api.semanticscholar.org/CorpusID:235436185}.

\bibitem[Beaumont(2021)]{beaumont-2021-img2dataset}
R.~Beaumont.
\newblock img2dataset: Easily turn large sets of image urls to an image dataset.
\newblock \url{https://github.com/rom1504/img2dataset}, 2021.

\bibitem[Beyer et~al.(2022)Beyer, Zhai, and Kolesnikov]{big_vision}
L.~Beyer, X.~Zhai, and A.~Kolesnikov.
\newblock Big vision.
\newblock \url{https://github.com/google-research/big_vision}, 2022.

\bibitem[Beyer et~al.(2023)Beyer, Wan, Madan, Pavetic, Steiner, Kolesnikov, Pinto, Bugliarello, Wang, Yu, Chen, and Zhai]{Beyer2023ASO}
L.~Beyer, B.~Wan, G.~Madan, F.~Pavetic, A.~Steiner, A.~Kolesnikov, A.~S. Pinto, E.~Bugliarello, X.~Wang, Q.~Yu, L.-C. Chen, and X.~Zhai.
\newblock A study of autoregressive decoders for multi-tasking in computer vision.
\newblock \emph{ArXiv}, abs/2303.17376, 2023.
\newblock URL \url{https://api.semanticscholar.org/CorpusID:257833784}.

\bibitem[Blier et~al.(2021)Blier, Tallec, and Ollivier]{blier2021learning}
L.~Blier, C.~Tallec, and Y.~Ollivier.
\newblock Learning successor states and goal-dependent values: A mathematical viewpoint.
\newblock \emph{arXiv preprint arXiv:2101.07123}, 2021.

\bibitem[Caron et~al.(2019)Caron, Bojanowski, Mairal, and Joulin]{Caron2019UnsupervisedPO}
M.~Caron, P.~Bojanowski, J.~Mairal, and A.~Joulin.
\newblock Unsupervised pre-training of image features on non-curated data.
\newblock \emph{2019 IEEE/CVF International Conference on Computer Vision (ICCV)}, pages 2959--2968, 2019.

\bibitem[Caron et~al.(2020)Caron, Misra, Mairal, Goyal, Bojanowski, and Joulin]{Caron2020UnsupervisedLO}
M.~Caron, I.~Misra, J.~Mairal, P.~Goyal, P.~Bojanowski, and A.~Joulin.
\newblock Unsupervised learning of visual features by contrasting cluster assignments.
\newblock \emph{ArXiv}, abs/2006.09882, 2020.
\newblock URL \url{https://api.semanticscholar.org/CorpusID:219721240}.

\bibitem[Caron et~al.(2021)Caron, Touvron, Misra, J'egou, Mairal, Bojanowski, and Joulin]{Caron2021EmergingPI}
M.~Caron, H.~Touvron, I.~Misra, H.~J'egou, J.~Mairal, P.~Bojanowski, and A.~Joulin.
\newblock Emerging properties in self-supervised vision transformers.
\newblock \emph{2021 IEEE/CVF International Conference on Computer Vision (ICCV)}, pages 9630--9640, 2021.

\bibitem[ChameleonTeam(2024)]{chameleonteam2024chameleonmixedmodalearlyfusionfoundation}
ChameleonTeam.
\newblock Chameleon: Mixed-modal early-fusion foundation models, 2024.
\newblock URL \url{https://arxiv.org/abs/2405.09818}.

\bibitem[Changpinyo et~al.(2021)Changpinyo, Sharma, Ding, and Soricut]{Changpinyo2021Conceptual1P}
S.~Changpinyo, P.~K. Sharma, N.~Ding, and R.~Soricut.
\newblock Conceptual 12m: Pushing web-scale image-text pre-training to recognize long-tail visual concepts.
\newblock \emph{2021 IEEE/CVF Conference on Computer Vision and Pattern Recognition (CVPR)}, pages 3557--3567, 2021.
\newblock URL \url{https://api.semanticscholar.org/CorpusID:231951742}.

\bibitem[Chen and Li(2020)]{chenintriguing}
T.~Chen and L.~Li.
\newblock Intriguing properties of contrastive losses.
\newblock \emph{CoRR}, abs/2011.02803, 2020.
\newblock URL \url{https://arxiv.org/abs/2011.02803}.

\bibitem[Chen et~al.(2020{\natexlab{a}})Chen, Kornblith, Norouzi, and Hinton]{Chen2020ASF}
T.~Chen, S.~Kornblith, M.~Norouzi, and G.~E. Hinton.
\newblock A simple framework for contrastive learning of visual representations.
\newblock \emph{ArXiv}, abs/2002.05709, 2020{\natexlab{a}}.

\bibitem[Chen et~al.(2020{\natexlab{b}})Chen, Kornblith, Swersky, Norouzi, and Hinton]{Chen2020BigSM}
T.~Chen, S.~Kornblith, K.~Swersky, M.~Norouzi, and G.~E. Hinton.
\newblock Big self-supervised models are strong semi-supervised learners.
\newblock \emph{CoRR}, abs/2006.10029, 2020{\natexlab{b}}.
\newblock URL \url{https://arxiv.org/abs/2006.10029}.

\bibitem[Chen and He(2020)]{Chen2020ExploringSS}
X.~Chen and K.~He.
\newblock Exploring simple siamese representation learning.
\newblock \emph{2021 IEEE/CVF Conference on Computer Vision and Pattern Recognition (CVPR)}, pages 15745--15753, 2020.
\newblock URL \url{https://api.semanticscholar.org/CorpusID:227118869}.

\bibitem[Chen et~al.(2021)Chen, Xie, and He]{Chen2021AnES}
X.~Chen, S.~Xie, and K.~He.
\newblock An empirical study of training self-supervised vision transformers.
\newblock \emph{2021 IEEE/CVF International Conference on Computer Vision (ICCV)}, pages 9620--9629, 2021.

\bibitem[Damen et~al.(2020)Damen, Doughty, Farinella, Fidler, Furnari, Kazakos, Moltisanti, Munro, Perrett, Price, and Wray]{Damen2020TheED}
D.~Damen, H.~Doughty, G.~M. Farinella, S.~Fidler, A.~Furnari, E.~Kazakos, D.~Moltisanti, J.~Munro, T.~Perrett, W.~Price, and M.~Wray.
\newblock The epic-kitchens dataset: Collection, challenges and baselines.
\newblock \emph{IEEE Transactions on Pattern Analysis and Machine Intelligence}, 43:\penalty0 4125--4141, 2020.
\newblock URL \url{https://api.semanticscholar.org/CorpusID:218470131}.

\bibitem[Darkhalil et~al.(2022)Darkhalil, Shan, Zhu, Ma, Kar, Higgins, Fidler, Fouhey, and Damen]{VISOR2022}
A.~Darkhalil, D.~Shan, B.~Zhu, J.~Ma, A.~Kar, R.~Higgins, S.~Fidler, D.~Fouhey, and D.~Damen.
\newblock Epic-kitchens visor benchmark: Video segmentations and object relations.
\newblock In \emph{Proceedings of the Neural Information Processing Systems (NeurIPS) Track on Datasets and Benchmarks}, 2022.

\bibitem[Dosovitskiy et~al.(2020)Dosovitskiy, Beyer, Kolesnikov, Weissenborn, Zhai, Unterthiner, Dehghani, Minderer, Heigold, Gelly, Uszkoreit, and Houlsby]{Dosovitskiy2020AnII}
A.~Dosovitskiy, L.~Beyer, A.~Kolesnikov, D.~Weissenborn, X.~Zhai, T.~Unterthiner, M.~Dehghani, M.~Minderer, G.~Heigold, S.~Gelly, J.~Uszkoreit, and N.~Houlsby.
\newblock An image is worth 16x16 words: Transformers for image recognition at scale.
\newblock \emph{ArXiv}, abs/2010.11929, 2020.
\newblock URL \url{https://api.semanticscholar.org/CorpusID:225039882}.

\bibitem[El-Nouby et~al.(2024)El-Nouby, Klein, Zhai, Bautista, Toshev, Shankar, Susskind, and Joulin]{elnouby2024scalablepretraininglargeautoregressive}
A.~El-Nouby, M.~Klein, S.~Zhai, M.~A. Bautista, A.~Toshev, V.~Shankar, J.~M. Susskind, and A.~Joulin.
\newblock Scalable pre-training of large autoregressive image models, 2024.
\newblock URL \url{https://arxiv.org/abs/2401.08541}.

\bibitem[Eysenbach et~al.(2022)Eysenbach, Zhang, Salakhutdinov, and Levine]{Eysenbach2022ContrastiveLA}
B.~Eysenbach, T.~Zhang, R.~Salakhutdinov, and S.~Levine.
\newblock Contrastive learning as goal-conditioned reinforcement learning.
\newblock \emph{ArXiv}, abs/2206.07568, 2022.
\newblock URL \url{https://api.semanticscholar.org/CorpusID:249674522}.

\bibitem[Fini et~al.(2023)Fini, Astolfi, Romero-Soriano, Verbeek, and Drozdzal]{fini2023improved}
E.~Fini, P.~Astolfi, A.~Romero-Soriano, J.~Verbeek, and M.~Drozdzal.
\newblock Improved baselines for vision-language pre-training.
\newblock \emph{Transactions on Machine Learning Research}, 2023.
\newblock ISSN 2835-8856.
\newblock URL \url{https://openreview.net/forum?id=a7nvXxNmdV}.
\newblock Featured Certification.

\bibitem[Fu et~al.(2024)Fu, Lian, Wang, Shi, Wang, Yala, Darrell, Efros, and Goldberg]{fu2024rethinking}
L.~Fu, L.~Lian, R.~Wang, B.~Shi, X.~Wang, A.~Yala, T.~Darrell, A.~A. Efros, and K.~Goldberg.
\newblock Rethinking patch dependence for masked autoencoders.
\newblock \emph{arXiv preprint arXiv:2401.14391}, 2024.

\bibitem[Ghasemipour et~al.(2020)Ghasemipour, Schuurmans, and Gu]{emaq}
S.~K.~S. Ghasemipour, D.~Schuurmans, and S.~S. Gu.
\newblock Emaq: Expected-max q-learning operator for simple yet effective offline and online {RL}.
\newblock \emph{CoRR}, abs/2007.11091, 2020.
\newblock URL \url{https://arxiv.org/abs/2007.11091}.

\bibitem[Grill et~al.(2020)Grill, Strub, Altch'e, Tallec, Richemond, Buchatskaya, Doersch, Pires, Guo, Azar, Piot, Kavukcuoglu, Munos, and Valko]{Grill2020BootstrapYO}
J.-B. Grill, F.~Strub, F.~Altch'e, C.~Tallec, P.~H. Richemond, E.~Buchatskaya, C.~Doersch, B.~{\'A}. Pires, Z.~D. Guo, M.~G. Azar, B.~Piot, K.~Kavukcuoglu, R.~Munos, and M.~Valko.
\newblock Bootstrap your own latent: A new approach to self-supervised learning.
\newblock \emph{ArXiv}, abs/2006.07733, 2020.

\bibitem[HaoChen and Ma(2023)]{haochen2023theoreticalstudyinductivebiases}
J.~Z. HaoChen and T.~Ma.
\newblock A theoretical study of inductive biases in contrastive learning, 2023.
\newblock URL \url{https://arxiv.org/abs/2211.14699}.

\bibitem[He et~al.(2019)He, Fan, Wu, Xie, and Girshick]{He2019MomentumCF}
K.~He, H.~Fan, Y.~Wu, S.~Xie, and R.~B. Girshick.
\newblock Momentum contrast for unsupervised visual representation learning.
\newblock \emph{2020 IEEE/CVF Conference on Computer Vision and Pattern Recognition (CVPR)}, pages 9726--9735, 2019.

\bibitem[He et~al.(2021)He, Chen, Xie, Li, Doll'ar, and Girshick]{He2021MaskedAA}
K.~He, X.~Chen, S.~Xie, Y.~Li, P.~Doll'ar, and R.~B. Girshick.
\newblock Masked autoencoders are scalable vision learners.
\newblock \emph{2022 IEEE/CVF Conference on Computer Vision and Pattern Recognition (CVPR)}, pages 15979--15988, 2021.
\newblock URL \url{https://api.semanticscholar.org/CorpusID:243985980}.

\bibitem[Iscen et~al.(2019)Iscen, Tolias, Avrithis, and Chum]{Iscen2019LabelPF}
A.~Iscen, G.~Tolias, Y.~Avrithis, and O.~Chum.
\newblock Label propagation for deep semi-supervised learning.
\newblock \emph{2019 IEEE/CVF Conference on Computer Vision and Pattern Recognition (CVPR)}, pages 5065--5074, 2019.
\newblock URL \url{https://api.semanticscholar.org/CorpusID:104291869}.

\bibitem[Jayaraman and Grauman(2018)]{jayaraman2018learning}
D.~Jayaraman and K.~Grauman.
\newblock Learning to look around: Intelligently exploring unseen environments for unknown tasks.
\newblock In \emph{Proceedings of the IEEE conference on computer vision and pattern recognition}, pages 1238--1247, 2018.

\bibitem[Jha et~al.(2024)Jha, Blaschko, Asano, and Tuytelaars]{jha2024commonstabilitymechanismselfsupervised}
A.~Jha, M.~B. Blaschko, Y.~M. Asano, and T.~Tuytelaars.
\newblock The common stability mechanism behind most self-supervised learning approaches, 2024.
\newblock URL \url{https://arxiv.org/abs/2402.14957}.

\bibitem[Jia et~al.(2021)Jia, Yang, Xia, Chen, Parekh, Pham, Le, Sung, Li, and Duerig]{Jia2021ScalingUV}
C.~Jia, Y.~Yang, Y.~Xia, Y.-T. Chen, Z.~Parekh, H.~Pham, Q.~V. Le, Y.-H. Sung, Z.~Li, and T.~Duerig.
\newblock Scaling up visual and vision-language representation learning with noisy text supervision.
\newblock \emph{ArXiv}, abs/2102.05918, 2021.
\newblock URL \url{https://api.semanticscholar.org/CorpusID:231879586}.

\bibitem[Jing et~al.(2021)Jing, Vincent, LeCun, and Tian]{Jing2021UnderstandingDC}
L.~Jing, P.~Vincent, Y.~LeCun, and Y.~Tian.
\newblock Understanding dimensional collapse in contrastive self-supervised learning.
\newblock \emph{ArXiv}, abs/2110.09348, 2021.
\newblock URL \url{https://api.semanticscholar.org/CorpusID:239016966}.

\bibitem[Johnson et~al.(2022)Johnson, Hanchi, and Maddison]{Johnson2022ContrastiveLC}
D.~D. Johnson, A.~E. Hanchi, and C.~J. Maddison.
\newblock Contrastive learning can find an optimal basis for approximately view-invariant functions.
\newblock \emph{ArXiv}, abs/2210.01883, 2022.
\newblock URL \url{https://api.semanticscholar.org/CorpusID:252715969}.

\bibitem[Karpathy and Li(2015)]{DBLP:conf/cvpr/KarpathyL15}
A.~Karpathy and F.~Li.
\newblock Deep visual-semantic alignments for generating image descriptions.
\newblock In \emph{{IEEE} Conference on Computer Vision and Pattern Recognition, {CVPR} 2015, Boston, MA, USA, June 7-12, 2015}, pages 3128--3137. {IEEE} Computer Society, 2015.
\newblock \doi{10.1109/CVPR.2015.7298932}.
\newblock URL \url{https://doi.org/10.1109/CVPR.2015.7298932}.

\bibitem[Kostrikov et~al.(2021)Kostrikov, Nair, and Levine]{kostrikov2021offline}
I.~Kostrikov, A.~Nair, and S.~Levine.
\newblock Offline reinforcement learning with implicit q-learning.
\newblock \emph{arXiv preprint arXiv:2110.06169}, 2021.

\bibitem[Lee et~al.(2013)]{lee2013pseudo}
D.-H. Lee et~al.
\newblock Pseudo-label: The simple and efficient semi-supervised learning method for deep neural networks.
\newblock In \emph{Workshop on challenges in representation learning, ICML}, volume~3, page 896. Atlanta, 2013.

\bibitem[Li et~al.(2022)Li, Efros, and Pathak]{SimSiamCollapse}
A.~C. Li, A.~A. Efros, and D.~Pathak.
\newblock Understanding collapse in non-contrastive siamese representation learning.
\newblock \emph{ECCV}, 2022.

\bibitem[Li et~al.(2021)Li, Liang, Zhao, Cui, Ouyang, Shao, Yu, and Yan]{Li2021SupervisionEE}
Y.~Li, F.~Liang, L.~Zhao, Y.~Cui, W.~Ouyang, J.~Shao, F.~Yu, and J.~Yan.
\newblock Supervision exists everywhere: A data efficient contrastive language-image pre-training paradigm.
\newblock \emph{ArXiv}, abs/2110.05208, 2021.
\newblock URL \url{https://api.semanticscholar.org/CorpusID:238582773}.

\bibitem[Lin et~al.(2014)Lin, Maire, Belongie, Hays, Perona, Ramanan, Doll{\'a}r, and Zitnick]{Lin2014MicrosoftCC}
T.-Y. Lin, M.~Maire, S.~J. Belongie, J.~Hays, P.~Perona, D.~Ramanan, P.~Doll{\'a}r, and C.~L. Zitnick.
\newblock Microsoft coco: Common objects in context.
\newblock In \emph{European Conference on Computer Vision}, 2014.
\newblock URL \url{https://api.semanticscholar.org/CorpusID:14113767}.

\bibitem[Mnih et~al.(2013)Mnih, Kavukcuoglu, Silver, Graves, Antonoglou, Wierstra, and Riedmiller]{Mnih2013PlayingAW}
V.~Mnih, K.~Kavukcuoglu, D.~Silver, A.~Graves, I.~Antonoglou, D.~Wierstra, and M.~A. Riedmiller.
\newblock Playing atari with deep reinforcement learning.
\newblock \emph{ArXiv}, abs/1312.5602, 2013.
\newblock URL \url{https://api.semanticscholar.org/CorpusID:15238391}.

\bibitem[Mu et~al.(2021)Mu, Kirillov, Wagner, and Xie]{Mu2021SLIPSM}
N.~Mu, A.~Kirillov, D.~A. Wagner, and S.~Xie.
\newblock Slip: Self-supervision meets language-image pre-training.
\newblock \emph{ArXiv}, abs/2112.12750, 2021.
\newblock URL \url{https://api.semanticscholar.org/CorpusID:245424883}.

\bibitem[Naeem et~al.(2023)Naeem, Xian, Zhai, Hoyer, Van~Gool, and Tombari]{naeem2023silc}
M.~F. Naeem, Y.~Xian, X.~Zhai, L.~Hoyer, L.~Van~Gool, and F.~Tombari.
\newblock Silc: Improving vision language pretraining with self-distillation.
\newblock \emph{arXiv preprint arXiv:2310.13355}, 2023.

\bibitem[Papanikolopoulos et~al.(1991)Papanikolopoulos, Khosla, and Kanade]{papanikolopoulos1991vision}
N.~Papanikolopoulos, P.~K. Khosla, and T.~Kanade.
\newblock Vision and control techniques for robotic visual tracking.
\newblock In \emph{ICRA}, pages 857--864, 1991.

\bibitem[Pathak et~al.(2016)Pathak, Kr{\"a}henb{\"u}hl, Donahue, Darrell, and Efros]{Pathak2016ContextEF}
D.~Pathak, P.~Kr{\"a}henb{\"u}hl, J.~Donahue, T.~Darrell, and A.~A. Efros.
\newblock Context encoders: Feature learning by inpainting.
\newblock \emph{2016 IEEE Conference on Computer Vision and Pattern Recognition (CVPR)}, pages 2536--2544, 2016.
\newblock URL \url{https://api.semanticscholar.org/CorpusID:2202933}.

\bibitem[Pham et~al.(2020)Pham, Xie, Dai, and Le]{Pham2020MetaPL}
H.~Pham, Q.~Xie, Z.~Dai, and Q.~V. Le.
\newblock Meta pseudo labels.
\newblock \emph{2021 IEEE/CVF Conference on Computer Vision and Pattern Recognition (CVPR)}, pages 11552--11563, 2020.

\bibitem[Radford et~al.(2021)Radford, Kim, Hallacy, Ramesh, Goh, Agarwal, Sastry, Askell, Mishkin, Clark, et~al.]{radford2021learning}
A.~Radford, J.~W. Kim, C.~Hallacy, A.~Ramesh, G.~Goh, S.~Agarwal, G.~Sastry, A.~Askell, P.~Mishkin, J.~Clark, et~al.
\newblock Learning transferable visual models from natural language supervision.
\newblock In \emph{International conference on machine learning}, pages 8748--8763. PMLR, 2021.

\bibitem[Rivlin and Rotstein(2000)]{rivlin2000control}
E.~Rivlin and H.~Rotstein.
\newblock Control of a camera for active vision: Foveal vision, smooth tracking and saccade.
\newblock \emph{International Journal of Computer Vision}, 39:\penalty0 81--96, 2000.

\bibitem[Rudner et~al.(2021)Rudner, Pong, McAllister, Gal, and Levine]{rudner2021outcome}
T.~G. Rudner, V.~Pong, R.~McAllister, Y.~Gal, and S.~Levine.
\newblock Outcome-driven reinforcement learning via variational inference.
\newblock \emph{Advances in Neural Information Processing Systems}, 34:\penalty0 13045--13058, 2021.

\bibitem[Russakovsky et~al.(2014)Russakovsky, Deng, Su, Krause, Satheesh, Ma, Huang, Karpathy, Khosla, Bernstein, Berg, and Fei-Fei]{Russakovsky2014ImageNetLS}
O.~Russakovsky, J.~Deng, H.~Su, J.~Krause, S.~Satheesh, S.~Ma, Z.~Huang, A.~Karpathy, A.~Khosla, M.~S. Bernstein, A.~C. Berg, and L.~Fei-Fei.
\newblock Imagenet large scale visual recognition challenge.
\newblock \emph{International Journal of Computer Vision}, 115:\penalty0 211 -- 252, 2014.
\newblock URL \url{https://api.semanticscholar.org/CorpusID:2930547}.

\bibitem[Sutton and Barto(2018)]{Sutton1998}
R.~S. Sutton and A.~G. Barto.
\newblock \emph{Reinforcement Learning: An Introduction}.
\newblock The MIT Press, second edition, 2018.
\newblock URL \url{http://incompleteideas.net/book/the-book-2nd.html}.

\bibitem[Sutton et~al.(2011)Sutton, Modayil, Delp, Degris, Pilarski, White, and Precup]{Sutton2011HordeAS}
R.~S. Sutton, J.~Modayil, M.~Delp, T.~Degris, P.~M. Pilarski, A.~White, and D.~Precup.
\newblock Horde: a scalable real-time architecture for learning knowledge from unsupervised sensorimotor interaction.
\newblock In \emph{Adaptive Agents and Multi-Agent Systems}, 2011.
\newblock URL \url{https://api.semanticscholar.org/CorpusID:13528549}.

\bibitem[Tian et~al.(2020)Tian, Sun, Poole, Krishnan, Schmid, and Isola]{tian2020makes}
Y.~Tian, C.~Sun, B.~Poole, D.~Krishnan, C.~Schmid, and P.~Isola.
\newblock What makes for good views for contrastive learning?
\newblock \emph{Advances in neural information processing systems}, 33:\penalty0 6827--6839, 2020.

\bibitem[van~den Oord et~al.(2018)van~den Oord, Li, and Vinyals]{Oord2018RepresentationLW}
A.~van~den Oord, Y.~Li, and O.~Vinyals.
\newblock Representation learning with contrastive predictive coding.
\newblock \emph{ArXiv}, abs/1807.03748, 2018.
\newblock URL \url{https://api.semanticscholar.org/CorpusID:49670925}.

\bibitem[Venkataramanan et~al.(2024)Venkataramanan, Rizve, Carreira, Asano, and Avrithis]{venkataramanan2023imagenet}
S.~Venkataramanan, M.~N. Rizve, J.~Carreira, Y.~M. Asano, and Y.~Avrithis.
\newblock Is imagenet worth 1 video? learning strong image encoders from 1 long unlabelled video.
\newblock In \emph{International Conference on Learning Representations}, 2024.

\bibitem[Watkins and Dayan(1992)]{watkins1992}
C.~Watkins and P.~Dayan.
\newblock Q-learning.
\newblock \emph{Machine Learning}, 8\penalty0 (3):\penalty0 279--292, 1992.
\newblock ISSN 1573-0565.
\newblock \doi{10.1007/BF00992698}.

\bibitem[Weers et~al.(2023)Weers, Shankar, Katharopoulos, Yang, and Gunter]{Weers_2023_CVPR}
F.~Weers, V.~Shankar, A.~Katharopoulos, Y.~Yang, and T.~Gunter.
\newblock Masked autoencoding does not help natural language supervision at scale.
\newblock In \emph{Proceedings of the IEEE/CVF Conference on Computer Vision and Pattern Recognition (CVPR)}, pages 23432--23444, June 2023.

\bibitem[Xie et~al.(2019)Xie, Dai, Hovy, Luong, and Le]{Xie2019UnsupervisedDA}
Q.~Xie, Z.~Dai, E.~H. Hovy, M.-T. Luong, and Q.~V. Le.
\newblock Unsupervised data augmentation for consistency training.
\newblock \emph{arXiv: Learning}, 2019.

\bibitem[Xie et~al.(2021)Xie, Zhang, Cao, Lin, Bao, Yao, Dai, and Hu]{Xie2021SimMIMAS}
Z.~Xie, Z.~Zhang, Y.~Cao, Y.~Lin, J.~Bao, Z.~Yao, Q.~Dai, and H.~Hu.
\newblock Simmim: a simple framework for masked image modeling.
\newblock \emph{2022 IEEE/CVF Conference on Computer Vision and Pattern Recognition (CVPR)}, pages 9643--9653, 2021.
\newblock URL \url{https://api.semanticscholar.org/CorpusID:244346275}.

\bibitem[Yang et~al.(2023)Yang, Song, King, and Xu]{semisupervised}
X.~Yang, Z.~Song, I.~King, and Z.~Xu.
\newblock A survey on deep semi-supervised learning.
\newblock \emph{IEEE Transactions on Knowledge and Data Engineering}, 35\penalty0 (9):\penalty0 8934--8954, 2023.
\newblock \doi{10.1109/TKDE.2022.3220219}.

\bibitem[Yu et~al.(2022)Yu, Wang, Vasudevan, Yeung, Seyedhosseini, and Wu]{Yu2022CoCaCC}
J.~Yu, Z.~Wang, V.~Vasudevan, L.~Yeung, M.~Seyedhosseini, and Y.~Wu.
\newblock Coca: Contrastive captioners are image-text foundation models.
\newblock \emph{Trans. Mach. Learn. Res.}, 2022, 2022.
\newblock URL \url{https://api.semanticscholar.org/CorpusID:248512473}.

\bibitem[Zhai et~al.(2019{\natexlab{a}})Zhai, Oliver, Kolesnikov, and Beyer]{Zhai2019S4LSS}
X.~Zhai, A.~Oliver, A.~Kolesnikov, and L.~Beyer.
\newblock S4l: Self-supervised semi-supervised learning.
\newblock \emph{2019 IEEE/CVF International Conference on Computer Vision (ICCV)}, pages 1476--1485, 2019{\natexlab{a}}.

\bibitem[Zhai et~al.(2019{\natexlab{b}})Zhai, Puigcerver, Kolesnikov, Ruyssen, Riquelme, Lucic, Djolonga, Pinto, Neumann, Dosovitskiy, Beyer, Bachem, Tschannen, Michalski, Bousquet, Gelly, and Houlsby]{Zhai2019ALS}
X.~Zhai, J.~Puigcerver, A.~Kolesnikov, P.~Ruyssen, C.~Riquelme, M.~Lucic, J.~Djolonga, A.~S. Pinto, M.~Neumann, A.~Dosovitskiy, L.~Beyer, O.~Bachem, M.~Tschannen, M.~Michalski, O.~Bousquet, S.~Gelly, and N.~Houlsby.
\newblock A large-scale study of representation learning with the visual task adaptation benchmark.
\newblock \emph{arXiv: Computer Vision and Pattern Recognition}, 2019{\natexlab{b}}.
\newblock URL \url{https://api.semanticscholar.org/CorpusID:214317405}.

\bibitem[Zhai et~al.(2023)Zhai, Mustafa, Kolesnikov, and Beyer]{Zhai2023SigmoidLF}
X.~Zhai, B.~Mustafa, A.~Kolesnikov, and L.~Beyer.
\newblock Sigmoid loss for language image pre-training.
\newblock \emph{2023 IEEE/CVF International Conference on Computer Vision (ICCV)}, pages 11941--11952, 2023.
\newblock URL \url{https://api.semanticscholar.org/CorpusID:257767223}.

\bibitem[Ziebart et~al.(2008)Ziebart, Maas, Bagnell, Dey, et~al.]{ziebart2008maximum}
B.~D. Ziebart, A.~L. Maas, J.~A. Bagnell, A.~K. Dey, et~al.
\newblock Maximum entropy inverse reinforcement learning.
\newblock In \emph{Aaai}, volume~8, pages 1433--1438. Chicago, IL, USA, 2008.

\end{thebibliography}
